\documentclass{article}
\usepackage{iclr2027_conference,times}
\usepackage[T1]{fontenc}
\usepackage[utf8]{inputenc}
\usepackage{colortbl}

\usepackage{amsmath,amsfonts,bm}

\def\eqref#1{equation~\ref{#1}}

\def\1{\bm{1}}

\DeclareMathAlphabet{\mathsfit}{\encodingdefault}{\sfdefault}{m}{sl}
\SetMathAlphabet{\mathsfit}{bold}{\encodingdefault}{\sfdefault}{bx}{n}

\usepackage{amsmath,amssymb,amsthm}
\usepackage{booktabs}
\usepackage{multirow}
\usepackage{graphicx}
\usepackage{float}
\usepackage{wrapfig}
\usepackage{hyperref}
\usepackage{url}
\usepackage[misc]{ifsym}

\title{Images Talk Back: Guiding MMDiTs by \\ Suppressing Image-to-Text Feedback}
\author{Yucheng Yang\textsuperscript{1}, Xiaoxiao Sun\textsuperscript{1}, Shusheng Yang\textsuperscript{2}
\\ \textbf{Jiakang Chen\textsuperscript{3}, Serena Yeung-Levy\textsuperscript{1}$^{\textrm{\Letter}}$} \\
\textsuperscript{1} Stanford University ~~~ \textsuperscript{2} New York University ~~~ \textsuperscript{3} University of California, Berkeley}

\newcommand{\sfg}{SFG}
\newcommand{\isfg}{iSFG}
\newcommand{\sfgs}{SFG-S}
\newcommand{\sfgx}{SFG-X}
\newcommand{\cfg}{CFG}
\newcommand{\vhat}{\hat{v}}
\newcommand{\img}{i}
\newcommand{\txt}{t}
\newcommand{\good}[1]{\textbf{#1}}
\newcommand{\second}[1]{\underline{#1}}

\graphicspath{{figures/}}
\iclrfinalcopy

\begin{document}

\maketitle
\fancyhead{}
\renewcommand{\headrulewidth}{0pt}

\vspace{-2mm}
\begin{figure}[ht]
\begin{center}
\includegraphics[width=0.94\linewidth]{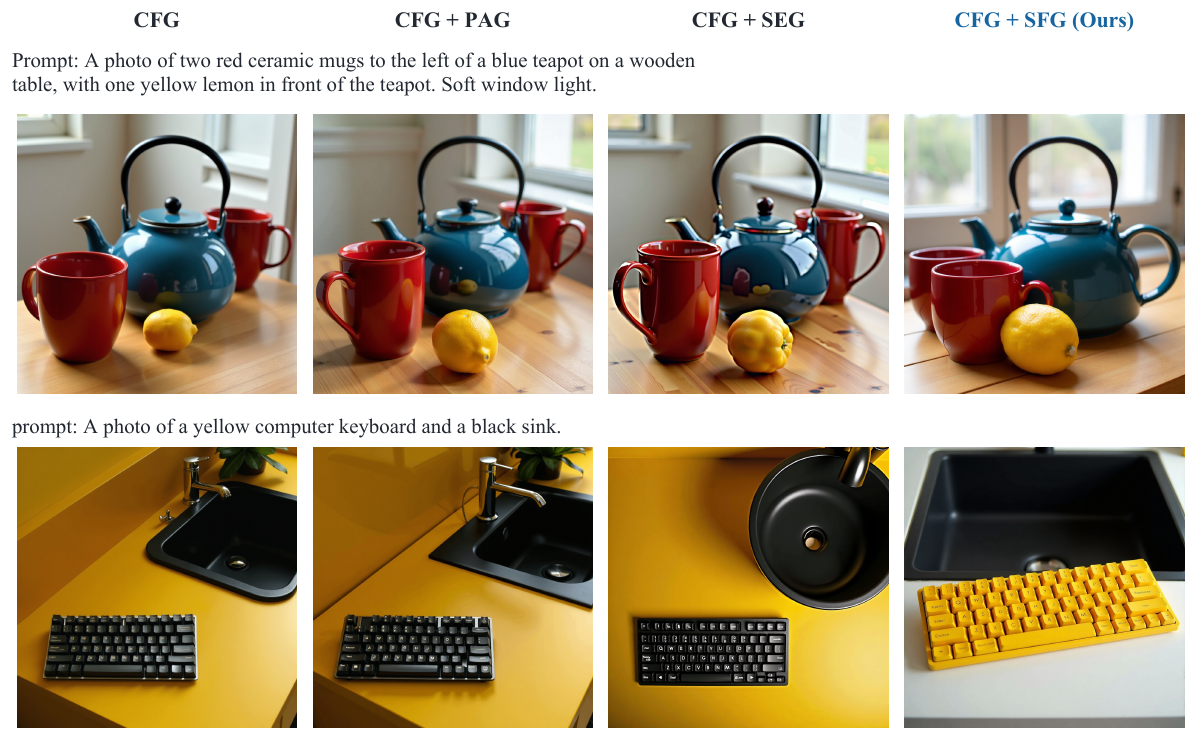}
\caption{\textbf{Matched qualitative comparisons on FLUX.} When dealing with more complex prompts, our method aligns better with the objects and attributes specified in the complex prompts.}
\label{fig:qualitative-intro}
\end{center}
\end{figure}

\begin{abstract}
Multimodal diffusion transformers (MMDiTs) allow the evolving image to update internal text representations through joint attention. These representations then guide subsequent image updates, raising a question: does this feedback help correct deviations from the prompt, or preserve them? We find that image-to-text (I2T) feedback can preserve the objects and attributes already represented in the evolving image, including errors. Transplanting I2T messages can rescue failures or disrupt successes. At matched intervention strengths, suppressing the influence of I2T feedback improves generation, whereas opposite controls degrade it. These findings motivate Semantic Feedback Guidance (\sfg), which suppresses the influence of I2T feedback by guiding away from an auxiliary branch that strengthens this feedback and weakens text-to-image conditioning. The in-place variant, \isfg{}, directly suppresses I2T feedback and strengthens text-to-image conditioning, requiring no extra model evaluations. Temporal interventions motivate running the \sfg{} branch during the first 20\% of image sampling steps, requiring about $1.2\times$ the base NFE without classifier-free guidance (CFG). Experiments across three image-generation backbones show gains in alignment and preference-model scores. On FLUX, adding \sfg{} to CFG raises GenEval from $0.6661$ to $0.7479$; on SD3.5M, \isfg{} adds $+1.33$ HPSv2.1 over CFG at zero extra NFE. The control also transfers to text-to-video and image-to-video generation, improving VBench scores. Code is available at \url{https://github.com/yyc1215lll-creator/SFG}.
\end{abstract}

\section{Introduction}
\label{sec:intro}

\begin{figure}[h]
\centering
\includegraphics[width=\linewidth]{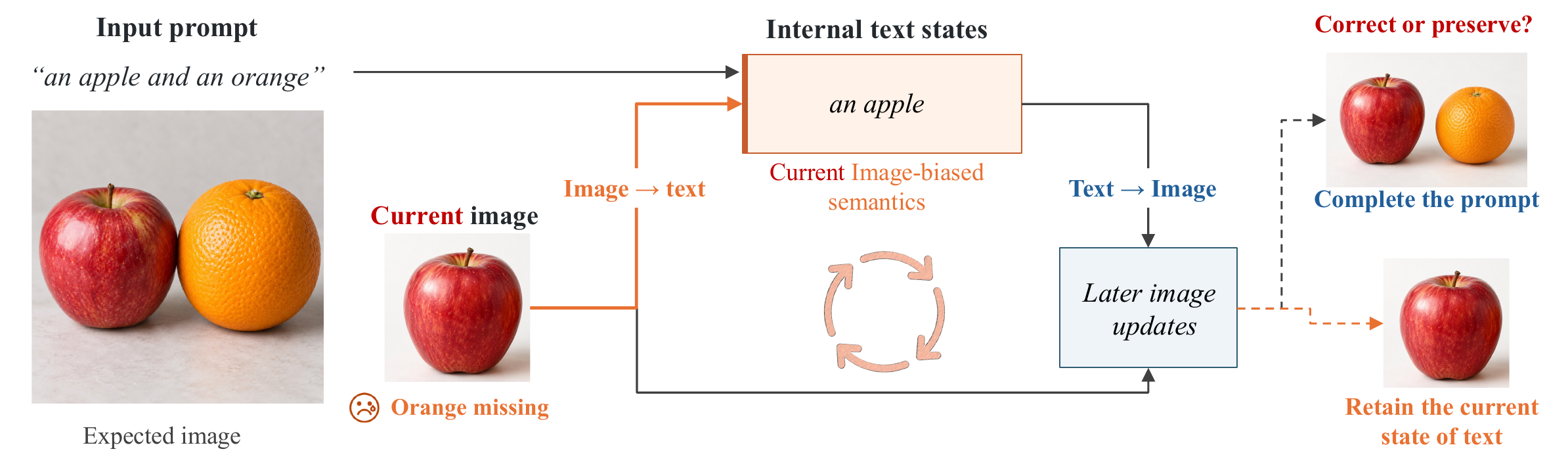}
\caption{\textbf{Image-to-text feedback.} The current image writes into internal text states through $B_{i\rightarrow t}$, which later blocks read to update the image. If the image misses an object, does this feedback help recover it or preserve the incomplete scene? Text labels and AI-generated fruit images illustrate schematic states and possible outcomes; the input prompt stays fixed.}
\label{fig:mechanism-overview}
\end{figure}

Text prompts specify what an image generator should produce, but in MMDiTs, the internal text stream also depends on what the model is drawing. Text queries read image keys and values before later blocks use the updated text stream to condition the image. This image-to-text-to-image loop underlies Stable Diffusion~3/3.5, FLUX, Qwen-Image, and HunyuanVideo \citep{esser2024sd3,flux2024,wu2025qwenimage,kong2024hunyuanvideo}. When the image departs from the prompt, does this feedback correct the departure or preserve the existing image content? For the prompt ``an apple and an orange,'' an image containing only an apple may write that incomplete scene back into the text stream. Figure~\ref{fig:mechanism-overview} illustrates the two possible effects: recovering the orange or reinforcing its omission. The answer determines whether early errors are corrected or persist through sampling.

Existing analyses describe the image-to-text path as a register for image content and object identity, or as the weaker cross-modal direction \citep{toker2025padding,li2026registers,shin2025mmdit}. They leave open how this content affects generation. We use \emph{image content} to mean the objects and attributes represented in the evolving image. Such content can conflict with the prompt; we test the effect of preserving it by intervening on this path.

To address these questions, we intervene on the image-to-text path and measure its effect on generation. We find that this feedback can preserve image content that conflicts with the prompt, and identify suppression as a useful direction. Specifically, we design two experiments to examine feedback content and local response direction. Message transplantation shows that feedback can rescue failures or disrupt successes. At the same failed state, feedback suppression is more aligned with the response to successful-message transplantation than with a norm-matched random control. Our first ablation tests finite interventions and shows that suppression improves generation, whereas the opposite control degrades it.

We construct a \emph{negative branch}, the reference prediction that guidance extrapolates away from, by increasing I2T feedback. Unlike SAG and PAG, which blur image regions or disrupt attention \citep{hong2023sag,ahn2024pag}, this branch preserves the attention computation and gives the current image more influence. Guiding away from it suppresses that influence. We call this control \sfgx{}. A complementary control, \sfgs{}, weakens text-to-image conditioning in the same negative branch, so guiding away strengthens conditioning. Together they form Semantic Feedback Guidance (\sfg{}). The in-place variant, \isfg{}, applies feedback suppression and stronger conditioning directly in the conditional forward, with no extra model evaluations.

Early message transplants have the greatest effect on the final outcome, consistent with object layout being largely fixed early in sampling, motivating a \sfg{} guidance window covering the first 20\% of image sampling steps. This costs about $1.2\times$ the base NFE without CFG, compared with $2\times$ for an auxiliary branch at every step. SFG corrections remain nearly orthogonal to CFG through sampling, and the two methods combine effectively. Experiments on SD3M, SD3.5M, and FLUX show gains in alignment, preference scores, and distribution quality; transfer to HunyuanVideo improves text-to-video and image-to-video VBench scores over the base model and CFG.

Our contributions are:
\begin{itemize}
\item \textbf{How I2T feedback preserves image content.} Bidirectional transplantation tests the content carried by I2T feedback. Local response analysis and matched two-sided interventions identify suppression as a useful direction on failed trajectories.
\item \textbf{A negative branch derived from feedback measurements.} SFG-X strengthens I2T feedback in its negative branch and guides against it. SFG combines this control with stronger T2I conditioning; temporal interventions motivate its guidance window.
\item \textbf{Branch and in-place controls.} iSFG applies the controls directly with zero extra NFE. Experiments evaluate alignment and preference gains, combination with CFG, and transfer to both video tasks.
\end{itemize}

\section{From mechanism analysis to signed guidance}
\label{sec:method}

\subsection{Background: guidance and writable MMDiT states}
\label{sec:prelim}

CFG extrapolates between conditional and unconditional predictions; training-free self-guidance extrapolates away from a perturbed prediction of the same model. The formulas and perturbation choices are summarized in App.~\ref{app:guidance-conventions}. We choose the perturbation by measuring a native feedback path.

\noindent\textbf{Writable MMDiT states.} Generation integrates a learned velocity field backward from noise along $x_t=(1-t)x_0+t\epsilon$, with $\epsilon\sim\mathcal N(0,I)$ \citep{esser2024sd3}. The encoded prompt $c=E(p)$ stays fixed. Conventional cross-attention and deep fusion only let image queries read prompt-derived features \citep{liu2024playground,tang2025deepfusion}. MMDiT instead maintains image ($\img$) and text ($\txt$) residual streams, both querying concatenated image--text keys through a joint softmax:
\begin{align}
  O_{\img} &= A_{\img\rightarrow\img}V_{\img} +
              \underbrace{A_{\txt\rightarrow\img}V_{\txt}}_{B_{\txt\rightarrow\img}}, &
  O_{\txt} &= A_{\txt\rightarrow\txt}V_{\txt} +
              \underbrace{A_{\img\rightarrow\txt}V_{\img}}_{B_{\img\rightarrow\txt}}.
  \label{eq:decomp}
\end{align}
We call $B_{\img\rightarrow\txt}$ the \emph{I2T message} and $B_{\txt\rightarrow\img}$ the \emph{text-to-image (T2I) message}. The I2T message writes image values into the text stream; the T2I message conditions the image on text. Later blocks can read this updated text back into the image, closing a feedback loop. Each denoiser call resets text from $E(p)$, so the stream changes with depth and the current $x_t$ without persisting across timesteps. We scale these messages after softmax, retaining joint normalization.

\subsection{Analysis: feedback content, direction, and timing}
\label{sec:motivation}\label{sec:motiv-feedback}\label{sec:feedback-sign}

\begin{figure}[h]
\centering
\includegraphics[width=\linewidth]{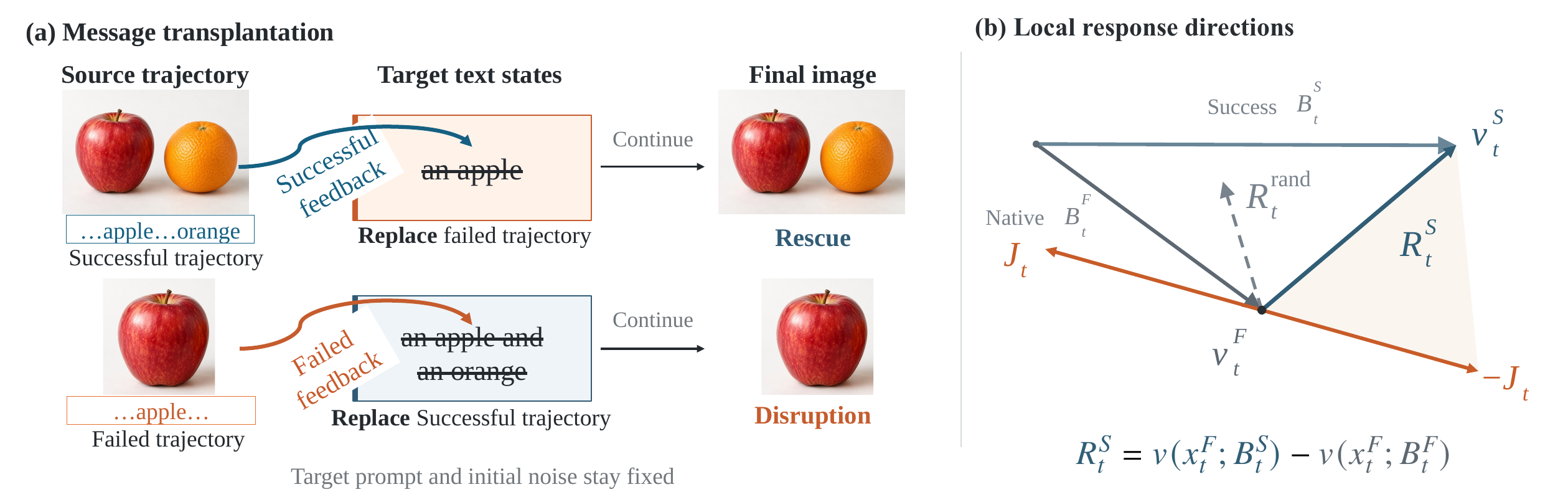}
\caption{\textbf{Message transplantation and local response directions.} (a) Transplant donor image-to-text messages into recipient text states, keeping the recipient prompt and initial noise fixed. Success-to-failure and failure-to-success transplants test rescue and disruption. (b) $J_t$ and $-J_t$ are opposite local responses; $R_t^S$ and $R_t^{\rm rand}$ are velocity changes from successful-donor and random-message interventions. Text labels, AI-generated fruit images, and vector angles are schematic.}
\label{fig:mechanism-protocol}
\end{figure}

We test what the I2T message carries, which intervention sign helps, and when it affects the final result. Figure~\ref{fig:mechanism-protocol} illustrates message transplantation and the local response directions. We then use those measurements to construct \sfg{} and \isfg{}.

\noindent\textbf{Paired trajectories.} We use 400 same-prompt pairs, each containing one successful and one failed trajectory from unmodified SD3.5M or FLUX runs, differing only in generation seed. Labels use GenEval \texttt{correct}, or T2I-CompBench scores $\geq0.75$ for success and $\leq0.45$ for failure; intermediate cases are excluded so that every pair has an unambiguous outcome label.

\noindent\textbf{Message transplantation.} We transplant the I2T message $B_{\img\rightarrow\txt}$. For success-to-failure transplantation, we cache the message of the successful donor by step and block and insert it after softmax and before output projection:
\begin{equation}
 B^{F\leftarrow S}_{t,\ell}(\lambda)
 =(1-\lambda)B^{F}_{t,\ell}+\lambda B^{S}_{t,\ell},
 \qquad \lambda\in[0,1],
 \label{eq:transplant}
\end{equation}

\begin{figure}[!t]
\centering
\begin{minipage}[t]{0.5\textwidth}
\vspace{0pt}
\centering
\includegraphics[width=\linewidth]{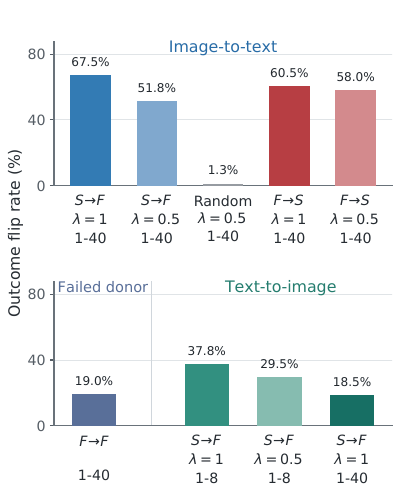}
\caption{\textbf{Message transplantation and controls.} $S/F$ denote successful/failed trajectories; arrows point from donor to recipient. Labels show donor mixing weight $\lambda$ and transplant steps. Full results: Table~\ref{tab:transplant-controls}.}
\label{fig:transplant}
\end{minipage}\hfill
\begin{minipage}[t]{0.485\textwidth}
\vspace{0pt}
\centering
\includegraphics[width=\linewidth]{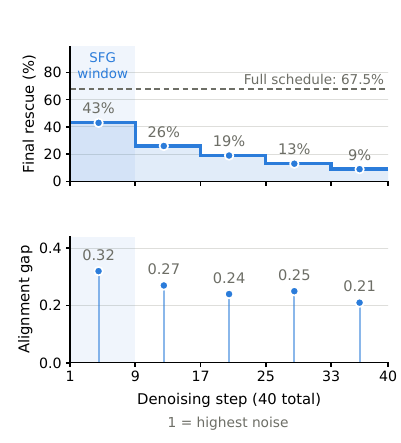}
\caption{\textbf{Timing and alignment} on 400 pairs across SD3.5M and FLUX. Top: rescue by transplant window; steps 1--8 of 40 recover about two-thirds of the full-schedule effect. Bottom: cosine gap over random, averaged within pairs, then within models, with equal model weights. Shading marks steps 1--8.}
\label{fig:temporal}
\end{minipage}
\end{figure}

The donor mixing weight $\lambda$ ranges from zero (retain) to one (fully replace). The prompt, seed, sampler, initial latent, and other activations of the recipient remain unchanged. We measure \emph{rescue} among initially failed recipients and \emph{disruption} among initially successful recipients: each counts a change to the opposite outcome. Norm-matched random messages rarely change the outcome (Figure~\ref{fig:transplant}).

We also test failed-donor and text-to-image transplants (Table~\ref{tab:transplant-controls}); both yield lower rescue rates than successful-donor image-to-text transplantation in the reported controls.

\noindent\textbf{Direction and timing.} On a separate fixed cohort of 400 success/failure pairs, we multiply the I2T message by $\alpha$ in every joint block. At each unmodified failed state, $J_t$ is the derivative of predicted velocity with respect to this multiplier at $\alpha=1$; $-J_t$ is the local response to reducing I2T feedback. Successful-donor transplantation changes the velocity by $R_t^S$, while a norm-matched random message perturbation changes it by $R_t^{\rm rand}$ (Figure~\ref{fig:mechanism-protocol}b). The \emph{cosine gap over random} compares their cosine similarities with $-J_t$, subtracting the random-control value from the successful-donor value. This gap is positive in both models across all time intervals (Figure~\ref{fig:temporal}). Transplanting during steps 1--8 of 40 recovers about two-thirds of the full-schedule rescues; late transplants rarely change the outcome. Definitions and aggregation details are in App.~\ref{app:jacobian}.

Appendix~\ref{app:object-probes} uses object-omission probes in the internal text stream to test whether errors persist once detected during sampling. Warnings usually persist without control; feedback suppression reduces this persistence and rescues many failures, with less frequent disruption of initially successful outputs.

\subsection{Method: SFG and iSFG}
\label{sec:control}

\sfg{} combines image-to-text feedback control (\sfgx{}) with text-to-image conditioning control (\sfgs{}). It implements both through an auxiliary prediction and an extrapolation step. The in-place variant, \isfg{}, applies the corresponding message scaling directly within conditional forward pass.

\noindent\textbf{Control targets and insertion point.} In every joint-attention block, we scale the I2T and T2I messages after softmax and value aggregation, before output projection. TACA and Attend-and-Excite strengthen T2I conditioning \citep{lv2025taca,chefer2023attendexcite}; SFG-X controls the I2T message, which neither method modifies. For a fixed block input, scaling preserves attention logits, joint normalization, and within-modality terms. Model weights stay frozen, but each forward recomputes activations, so earlier edits can affect later blocks through the updated text stream.

\noindent\textbf{SFG: auxiliary-branch guidance.} For edit strength $u<0$, the auxiliary branch increases image-to-text feedback and decreases text-to-image conditioning with tied magnitudes:
\begin{equation}
\begin{aligned}
O_{\txt}^{\rm pert}(u)
 &= A_{\txt\rightarrow\txt}V_{\txt}
    +(1-u)B_{\img\rightarrow\txt},\\
O_{\img}^{\rm pert}(u)
 &= A_{\img\rightarrow\img}V_{\img}
    +(1-|u|)B_{\txt\rightarrow\img}.
\end{aligned}
\label{eq:pathlinear}
\end{equation}
The first edit defines \sfgx{}, and the second defines \sfgs{}; \sfg{} applies both in the same auxiliary forward. To use either component alone, leave the other message at its original multiplier of one. Setting $u=0$ removes both edits. Text-to-image attenuation uses $|u|\leq1$; larger magnitudes reverse the sign of that message.

We call an unmodified conditional forward at the current latent the \emph{clean branch}. At the same $x_t$, $t$, and $c$, its prediction is $v_{\rm clean}=v_{\boldsymbol\theta}(x_t,t,c)$; $v_{\rm pert}$ is the auxiliary-branch prediction with the selected message scaling. With guidance strength $\omega\geq0$ and CFG scale $s\geq1$,
\begin{equation}
\begin{aligned}
\vhat=v_{\rm clean}
 &+\omega\left(v_{\rm clean}-v_{\rm pert}\right)\\
 &+(s-1)\left[v_{\rm clean}
 -v_{\boldsymbol\theta}(x_t,t,\varnothing)\right].
\end{aligned}
\label{eq:explicit}
\end{equation}
Here $u$ sets the message multipliers, $\omega$ is the SFG guidance scale, and $s$ is the CFG scale. SFG-X constructs its negative branch by strengthening I2T feedback ($1-u>1$), rather than degrading the model. This strengthening of a native pathway takes the opposite approach to guidance methods such as PAG~\citep{ahn2024pag}, SEG~\citep{hong2024seg}, and PG~\citep{chen2026guidedbridge}, which construct auxiliary branches by degrading the model. Guiding away suppresses this feedback, as motivated by Section~\ref{sec:motivation}, while SFG-S strengthens conditioning by guiding away from its attenuation. The sampler advances one trajectory using $\vhat$. At $s=1$, it omits the unconditional evaluation; otherwise, that prediction remains unmodified.

\noindent\textbf{iSFG: in-place control.} \isfg{} edits the ordinary conditional forward directly:
\begin{equation}
\begin{aligned}
O_{\txt}^{\rm iSFG}
 &= A_{\txt\rightarrow\txt}V_{\txt}
    +(1+\gamma)B_{\img\rightarrow\txt},\\
O_{\img}^{\rm iSFG}
 &= A_{\img\rightarrow\img}V_{\img}
    +(1+\gamma_{\rm S})B_{\txt\rightarrow\img}.
\end{aligned}
\label{eq:isfg}
\end{equation}
The I2T multiplier is $1+\gamma$ with $\gamma<0$; the T2I multiplier is $1+\gamma_{\rm S}$ with $\gamma_{\rm S}>0$. Thus, $u<0$ strengthens feedback in the negative branch, so guiding away suppresses its influence; $\gamma<0$ applies suppression in place. The component ablation attributes most of the gain of SFG to feedback control (Section~\ref{sec:component-window-ablation}, Figure~\ref{fig:component-window-ablation}a). Let $v_{\rm ip}$ be the resulting conditional prediction. Without CFG, the sampler uses $v_{\rm ip}$ directly; with CFG, it uses $s v_{\rm ip}-(s-1)v_{\boldsymbol\theta}(x_t,t,\varnothing)$. Only the conditional branch is edited. These edit strengths are independent of $\omega$; relations between matched branch and in-place controls are given in App.~\ref{app:path-theory}.

\noindent\textbf{Sampling schedule and computational cost.} \sfg{} uses approximately $1.2K$ model evaluations without CFG and $2.2K$ with CFG, activating its auxiliary branch during the first 20\% of $K$ image sampling steps. \isfg{} adds no model evaluations and has minimal runtime overhead (App.~\ref{app:impl}).

\suppressfloats[t]
\section{Experiments: evaluating mechanism-grounded guidance}
\label{sec:exp}

\noindent\textbf{Tasks, datasets, and metrics.} We evaluate text-to-image generation on SD3M, SD3.5M, and FLUX at $1024^2$, and text-to-video and image-to-video generation on HunyuanVideo \citep{kong2024hunyuanvideo}. For images, we evaluate compositional text-image alignment on GenEval \citep{ghosh2023geneval} and T2I-CompBench++ \citep{huang2025t2icompbenchpp}. We report HPSv2.1 \citep{wu2023hpsv2} and ImageReward \citep{xu2023imagereward} on the same 3,200 images: our evaluation set contains 800 prompts in total across four styles, with four seeds per prompt. We also report FID on COCO for SD3M. For video, we use VBench \citep{huang2024vbench}, evaluating seven text-to-video dimensions and eight image-to-video dimensions, and report the aggregate total score for each task.

\noindent\textbf{Baselines.} Image baselines include PAG \citep{ahn2024pag}, SEG \citep{hong2024seg}, S$^2$-Guidance \citep{chen2026s2guidance}, and TACA \citep{lv2025taca}, with comparisons both without and on top of \cfg{}. TACA strengthens text-to-image conditioning, the reverse information-flow direction from \sfgx{}, making it a close attention-rebalancing control for \sfgs{}. We also include Prompt Reinjection, which restores shallow text features \citep{yao2026promptreinjection}. Video baselines include \cfg{}, S$^2$-Guidance, and CFG-Zero$^\star$ \citep{fan2025cfgzero}.

\noindent\textbf{Evaluation protocol.} Within each configuration, methods share prompts and seeds. GenEval comparisons and guidance-scale sweeps use 553 prompts with six seeds per prompt, giving 3,318 images per configuration. For image benchmark comparisons, the \sfg{} branch runs during the first $20\%$ of sampling steps. Sample counts and generation settings appear in App.~\ref{app:hparams}, baseline weight sweeps in App.~\ref{app:baseline-sweeps}, and single- and two-object probe diagnostics with separate intervention schedules in App.~\ref{app:object-probes}. In Tables~\ref{tab:geneval-nocfg}--\ref{tab:i2v-headline}, higher scores are better except for FID; bold and underlined values mark the best and second-best results within each model.

\subsection{Image generation}
\label{sec:image-generation}

\noindent\textbf{Guidance without CFG.} At $\mathrm{cfg}=1$, \sfg{} improves GenEval and T2I-CompBench++ on both backbones (Table~\ref{tab:geneval-nocfg}). On SD3M, the scores rise from $0.3294$ to $0.5811$ and from $0.3230$ to $0.4760$, respectively; on FLUX, they rise from $0.3604$ to $0.6853$ and from $0.3254$ to $0.4727$. GenEval gains over the strongest competing method are $0.0207$ on SD3M (PAG) and $0.1305$ on FLUX (SEG). Feedback control drives most of the improvement over the base model: \sfgx{} alone exceeds every baseline on all reported metrics and backbones, while \sfgs{} alone trails PAG and SEG on GenEval. SFG also reduces SD3M COCO FID from $41.76$ to $24.21$, below the best competing score of $26.00$ and the $32.49$ obtained with CFG at $s=7.5$, without evaluating an unconditional branch.

\begin{table}[t]
\caption{\textbf{Text-to-image generation without CFG on SD3M and FLUX.} All methods use $\mathrm{cfg}=1$. GenEval and T2I-CompBench++ are averaged over six and eight categories, respectively. FID uses 5k COCO images on SD3M; CFG~\citep{ho2022classifierfree} at $s=7.5$ gives $32.49$ for reference. Dashes denote unavailable results.}
\label{tab:geneval-nocfg}\label{tab:t2icomp}\label{tab:fid}
\centering
\footnotesize\fontsize{8}{9.5}\selectfont
\setlength{\tabcolsep}{3.5pt}
\renewcommand{\arraystretch}{1.08}
\def\tableonebody{
\toprule
 & \multicolumn{3}{c}{SD3M} & \multicolumn{2}{c}{FLUX} \\
\cmidrule(lr){2-4}\cmidrule(lr){5-6}
Method & GenEval $\uparrow$ & \shortstack{T2I-Comp\\Bench++ $\uparrow$} & FID $\downarrow$ & GenEval $\uparrow$ & \shortstack{T2I-Comp\\Bench++ $\uparrow$} \\
\midrule
Base & 0.3294 & 0.3230 & 41.76 & 0.3604 & 0.3254 \\
S$^2$-Guidance \citep{chen2026s2guidance} & 0.4866 & 0.3667 & 26.00 & 0.4990 & 0.3900 \\
SEG \citep{hong2024seg} & 0.5322 & 0.4331 & 28.19 & 0.5548 & 0.4505 \\
PAG \citep{ahn2024pag} & 0.5604 & 0.4278 & 28.53 & 0.4866 & 0.3843 \\
\rowcolor[gray]{0.94}
\sfgs{} (ours) & 0.4678 & 0.4249 & 40.19 & 0.4471 & 0.3961 \\
\rowcolor[gray]{0.94}
\sfgx{} (ours) & \second{0.5621} & \second{0.4505} & \second{24.90} & \second{0.6277} & \second{0.4515} \\
\rowcolor[gray]{0.94}
\sfg{} (ours) & \good{0.5811} & \good{0.4760} & \good{24.21} & \good{0.6853} & \good{0.4727} \\
\bottomrule
}
\sbox0{\begin{tabular}{lrrrrr}\tableonebody\end{tabular}}
\setlength{\tabcolsep}{\dimexpr\tabcolsep+(0.95\linewidth-\wd0)/12\relax}
\begin{tabular}{lrrrrr}\tableonebody\end{tabular}
\end{table}

\noindent\textbf{Guidance on top of CFG.} Adding \sfg{} to CFG achieves the highest GenEval, HPSv2.1, and ImageReward scores on both backbones (Table~\ref{tab:geneval-cfg}). GenEval rises from $0.6890$ to $0.7110$ on SD3.5M and from $0.6661$ to $0.7479$ on FLUX, where counting rises from $0.6458$ to $0.8167$ and position from $0.1850$ to $0.3033$; HPSv2.1 increases by $0.85$ and $0.70$. In contrast, PAG, SEG, and S$^2$-Guidance do not raise overall GenEval above CFG on either backbone, and TACA, which only strengthens text-to-image conditioning, reaches $0.6960$ on SD3.5M. No baseline isolates the I2T message.

\begin{table}[t]
\caption{\textbf{Text-to-image generation with CFG on SD3.5M and FLUX.} Each additional guidance method is combined with CFG. Count., Pos., and C.-Attr. denote counting, position, and color-attribute GenEval scores, selected to show categories with larger SFG gains; Overall averages all six categories. HPSv2.1 averages the four style scores; ImageReward reports the mean over the same images, with identical prompts and seeds for every method.}
\label{tab:geneval-cfg}\label{tab:hps}
\centering
\footnotesize\fontsize{8}{9.5}\selectfont
\setlength{\tabcolsep}{2.5pt}
\begin{tabular}{@{}llrrrrrr@{}}
\toprule
& & \multicolumn{4}{c}{GenEval $\uparrow$} & HPS $\uparrow$ & ImageReward $\uparrow$ \\
\cmidrule(lr){3-6}\cmidrule(lr){7-7}\cmidrule(lr){8-8}
Model & Method & Count. & Pos. & C.-Attr. & Overall & Avg. & Mean \\
\midrule
\multirow{6}{*}{SD3.5M} & \cfg{}~\citep{ho2022classifierfree} & 0.6458 & \second{0.2150} & 0.5767 & 0.6890 & 31.65 & 1.1102 \\
 & \cfg+PAG \citep{ahn2024pag} & 0.6458 & 0.2017 & 0.5233 & 0.6807 & 31.51 & 1.1038 \\
 & \cfg+SEG \citep{hong2024seg} & 0.6646 & 0.2017 & \second{0.6017} & 0.6890 & \second{32.11} & 1.1180 \\
 & \cfg+S$^2$ \citep{chen2026s2guidance} & 0.6458 & 0.1967 & 0.5717 & 0.6811 & 31.71 & 1.1152 \\
 & \cfg+TACA \citep{lv2025taca} & \second{0.6771} & 0.1900 & 0.5950 & \second{0.6960} & 31.58 & \second{1.1298} \\
 & \cfg+\sfg{} (ours) & \good{0.6875} & \good{0.2250} & \good{0.6117} & \good{0.7110} & \good{32.50} & \good{1.1587} \\
\midrule
\multirow{6}{*}{FLUX} & \cfg{}~\citep{ho2022classifierfree} & 0.6458 & 0.1850 & 0.5233 & 0.6661 & 30.96 & 0.9837 \\
 & \cfg+PAG \citep{ahn2024pag} & 0.5792 & 0.1900 & \second{0.5300} & 0.6628 & 30.83 & 0.9765 \\
 & \cfg+SEG \citep{hong2024seg} & 0.6271 & 0.1900 & 0.4650 & 0.6500 & 31.10 & 1.0201 \\
 & \cfg+S$^2$ \citep{chen2026s2guidance} & 0.6458 & 0.1733 & 0.5117 & 0.6605 & 31.02 & 0.9912 \\
 & \cfg+Prompt Reinjection \citep{yao2026promptreinjection} & \second{0.7500} & \second{0.2500} & 0.5233 & \second{0.6975} & \second{31.25} & \second{1.0211} \\
 & \cfg+\sfg{} (ours) & \good{0.8167} & \good{0.3033} & \good{0.5933} & \good{0.7479} & \good{31.66} & \good{1.0436} \\
\bottomrule
\end{tabular}
\end{table}

\noindent\textbf{Zero-extra-NFE guidance with iSFG.} Without an auxiliary branch, \isfg{} improves CFG on all metrics (Table~\ref{tab:isfg}). GenEval rises from $0.6890$ to $0.7047$ on SD3.5M and from $0.6661$ to $0.6868$ on FLUX; HPSv2.1 increases by $1.33$ and $0.32$, and ImageReward by $0.0748$ and $0.0532$. Among reported categories, counting improves most on SD3.5M, rising from $0.6458$ to $0.6813$. Compared with SFG, iSFG gives smaller overall GenEval gains but higher preference scores on SD3.5M ($32.98$ vs.\ $32.50$ HPSv2.1 and $1.1850$ vs.\ $1.1587$ ImageReward), at zero extra NFE.

\begin{table}[t]
\caption{\textbf{Improving CFG with iSFG at zero additional NFE on SD3.5M and FLUX.} iSFG edits the conditional forward pass without increasing model evaluations relative to CFG. Count., Pos., and C.-Attr. are selected GenEval categories with larger gains: counting, position, and color attribute; Overall averages six categories. HPSv2.1 averages the four style scores; ImageReward reports the mean over the same images.}
\label{tab:isfg}
\centering
\footnotesize\fontsize{8}{9.5}\selectfont
\setlength{\tabcolsep}{3.5pt}
\begin{tabular}{@{}llrrrrrr@{}}
\toprule
& & \multicolumn{4}{c}{GenEval $\uparrow$} & HPSv2.1 $\uparrow$ & ImageReward $\uparrow$ \\
\cmidrule(lr){3-6}\cmidrule(lr){7-7}\cmidrule(lr){8-8}
Model & Method & Count. & Pos. & C.-Attr. & Overall & Avg. & Mean \\
\midrule
\multirow{2}{*}{SD3.5M} & \cfg{}~\citep{ho2022classifierfree} & \second{0.6458} & \second{0.2150} & \second{0.5767} & \second{0.6890} & \second{31.65} & \second{1.1102} \\
 & \cfg+\isfg{} (ours) & \good{0.6813} & \good{0.2367} & \good{0.5950} & \good{0.7047} & \good{32.98} & \good{1.1850} \\
\midrule
\multirow{2}{*}{FLUX} & \cfg{}~\citep{ho2022classifierfree} & \second{0.6458} & \second{0.1850} & \second{0.5233} & \second{0.6661} & \second{30.96} & \second{0.9837} \\
 & \cfg+\isfg{} (ours) & \good{0.6563} & \good{0.2200} & \good{0.5417} & \good{0.6868} & \good{31.28} & \good{1.0369} \\
\bottomrule
\end{tabular}
\end{table}

\subsection{Transfer to video generation}
\label{sec:video-generation}

\noindent\textbf{Text-to-video generation.}
\label{sec:t2v}
In both video tasks, SFG runs with $\mathrm{cfg}=1$ and requires no training. For text-to-video, seven active steps of 25 require 32 branch predictions versus 50 for full-schedule CFG and 39 for CFG limited to its first 14 steps. SFG reaches $84.25$ on seven VBench dimensions, exceeding all compared methods, including the base model ($82.58$; Table~\ref{tab:vbench}). Its $+1.67$ gain over the base model is more than four times that of S$^2$-Guidance ($+0.38$), the other self-guidance baseline. It leads in subject consistency and imaging quality ($93.71$ and $66.61$, versus at most $93.08$ and $64.94$ under CFG). CFG retains higher background consistency and aesthetics; S$^2$-Guidance has the highest dynamic degree. Unlike CFG, which lowers temporal flickering and motion smoothness below the base model, SFG keeps both at or near base levels ($98.18$ and $99.36$). It also raises aesthetic quality from $50.88$ to $56.00$, recovering most of the gain from CFG.

\begin{table}[t]
\caption{\textbf{Text-to-video generation with different guidance methods on HunyuanVideo.} We report seven VBench quality dimensions and the aggregate score (Total). SFG uses $\mathrm{cfg}=1$ without an unconditional branch; the two CFG rows differ in their guidance window. Cons., Flicker, and Smooth. denote consistency, temporal flickering, and motion smoothness.}
\label{tab:vbench}
\centering
\footnotesize\fontsize{8}{9.5}\selectfont
\setlength{\tabcolsep}{1.2pt}
\begin{tabular}{lcccccccc}
\toprule
& Total & Subject & Background & Temporal & Motion & Dynamic & Aesthetic & Imaging \\
Method & Score & Cons. & Cons. & Flicker & Smooth. & Degree & Quality & Quality \\
\midrule
Base (no guidance)      & 82.58 & 92.11 & 91.98 & \second{98.01} & \good{99.39} & \second{97.92} & 50.88 & 63.65 \\
\cfg{} (all steps)~\citep{ho2022classifierfree} & 83.52 & 92.79 & \second{92.65} & 97.20 & 98.78 & 96.53 & \good{58.97} & \second{64.94} \\
\cfg{} (first 14 steps)~\citep{ho2022classifierfree} & \second{83.58} & \second{93.08} & \good{92.67} & 97.38 & 99.00 & 96.53 & 57.81 & 64.90 \\
S$^2$-Guidance~\citep{chen2026s2guidance} & 82.96 & 92.35 & 92.43 & 97.49 & 99.35 & \good{98.61} & 52.69 & 64.62 \\
CFG-Zero$^\star$~\citep{fan2025cfgzero} & 83.41 & 92.74 & 92.38 & 96.97 & 98.86 & \second{97.92} & \second{58.41} & 64.86 \\
\sfg{} (ours)           & \good{84.25} & \good{93.71} & 92.37 & \good{98.18} & \second{99.36} & \second{97.92} & 56.00 & \good{66.61} \\
\bottomrule
\end{tabular}
\end{table}

\noindent\textbf{Image-to-video generation.}
\label{sec:i2v}
For image-to-video, reference-image and text tokens jointly form the conditioning stream. SFG applies unchanged to this entire stream, suppressing feedback from the evolving video and strengthening conditioning, and thus generalizes from text-only to multimodal conditioning. SFG scores $90.04$, versus $89.43$ for CFG and $88.50$ for the base model, and leads in eight of the nine reported columns (Table~\ref{tab:i2v-headline}). Dynamic degree rises from $66.67$ under CFG to $70.44$, although it remains below the base model and S$^2$-Guidance. SFG also achieves the highest consistency with the reference image. CFG-Zero$^\star$ falls below the base model ($87.49$).

\begin{table}[!htbp]
\caption{\textbf{Image-to-video generation with different guidance methods on HunyuanVideo.} We report eight VBench dimensions and the aggregate score (Total), covering reference-image consistency (I2V Subj./Bg.) and video quality (e.g., Subject/Background Cons.).}
\label{tab:i2v-headline}
\centering
\footnotesize\fontsize{8}{9.5}\selectfont
\setlength{\tabcolsep}{2pt}
\begin{tabular}{lccccccccc}
\toprule
& Total & I2V & I2V & Subject & Background & Motion & Dynamic & Aesthetic & Imaging \\
Method & Score & Subj. & Bg. & Cons. & Cons. & Smooth. & Degree & Quality & Quality \\
\midrule
Base (no guidance) & 88.50 & 96.51 & 97.92 & 89.83 & 91.73 & 98.99 & 85.37 & 57.62 & 67.88 \\
\cfg{}~\citep{ho2022classifierfree} & \second{89.43} & \second{97.82} & \second{98.59} & \second{92.88} & \second{94.43} & \second{99.10} & 66.67 & \second{60.44} & \second{70.24} \\
S$^2$-Guidance~\citep{chen2026s2guidance} & 88.66 & 96.63 & 97.98 & 90.01 & 91.74 & 98.93 & \good{86.59} & 58.12 & 67.90 \\
CFG-Zero$^\star$~\citep{fan2025cfgzero} & 87.49 & 95.79 & 97.19 & 86.59 & 90.61 & 98.87 & \second{85.77} & 57.23 & 67.74 \\
\sfg{} (ours)      & \good{90.04} & \good{98.16} & \good{99.06} & \good{93.04} & \good{94.57} & \good{99.50} & 70.44 & \good{60.56} & \good{70.44} \\
\bottomrule
\end{tabular}
\end{table}

\subsection{Ablation studies}
\label{sec:component-window-ablation}

\begin{wrapfigure}{r}{0.5\linewidth}
\vspace{-12pt}
\centering
\includegraphics[width=\linewidth]{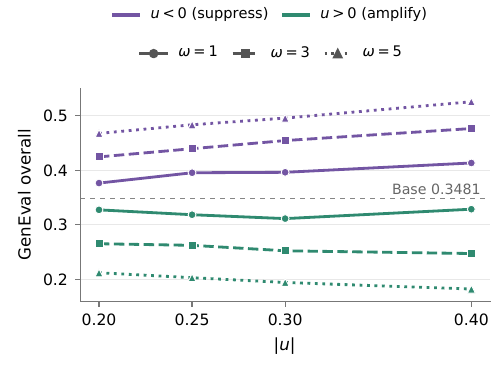}
\caption{\textbf{Validation of the \sfgx{} guidance direction.} GenEval on SD3.5M ($\mathrm{cfg}=1$), comparing \sfgx{} ($u<0$) with the opposite-sign branch ($u>0$) at matched $|u|$ and $\omega$. Suppression/amplification labels denote the net guidance effect. Base marks the unguided score.}
\label{fig:signcheck}
\vspace{-10pt}
\end{wrapfigure}

\noindent\textbf{Guidance direction.} Figure~\ref{fig:signcheck} tests the local response direction with finite edits on SD3.5M at $\mathrm{cfg}=1$. We compare both signs at matched magnitudes $|u|\in\{0.20,0.25,0.30,0.40\}$ and scales $\omega\in\{1,3,5\}$. All twelve suppression settings ($u<0$) improve on the base GenEval score of $0.3481$; all twelve opposite-sign settings reduce it. Larger scales widen the separation, with the strongest opposite-sign setting falling to roughly half the base score.

\begin{figure}[t]
\centering
\includegraphics[width=\linewidth]{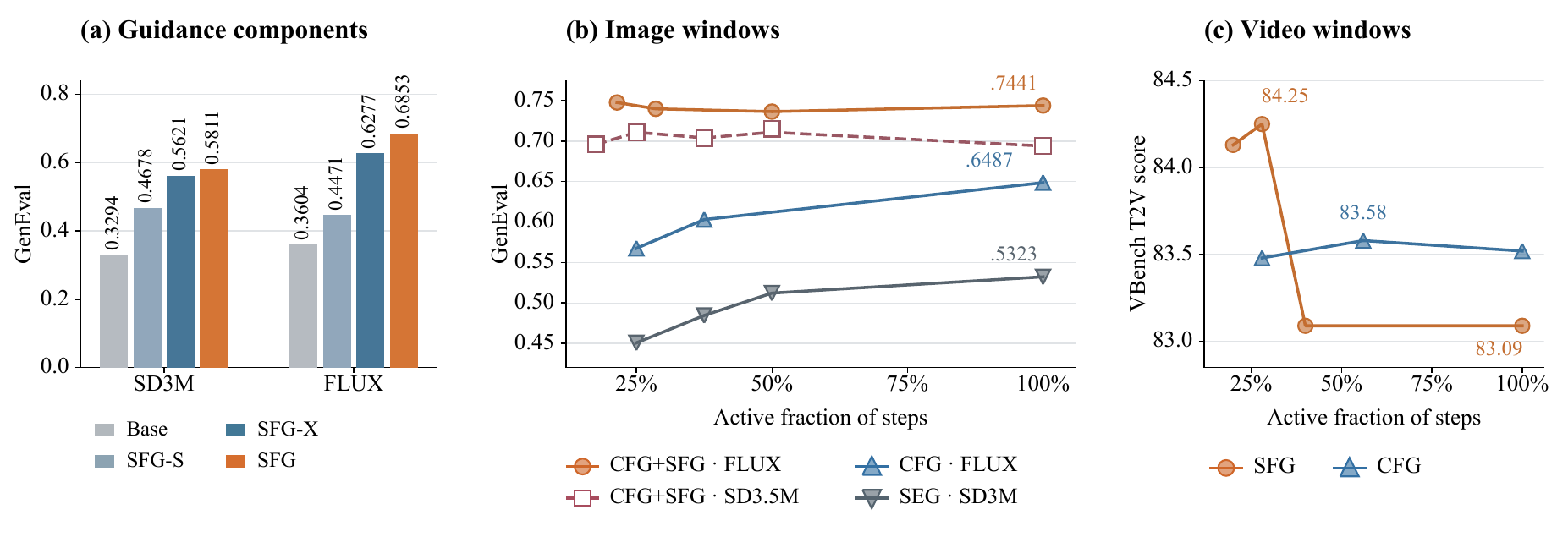}
\caption{\textbf{Guidance components and guidance windows.} (a) GenEval at the Table~\ref{tab:geneval-nocfg} operating points. (b) Both FLUX curves use 28 steps; SD3.5M CFG+SFG and SD3M SEG use 40. FLUX CFG returns to scale 1 outside its guidance window. (c) HunyuanVideo T2V scores with 25 total sampling steps. SFG is active for the first 5, 7, 10, or 25 steps; CFG is active for the first 7, 14, or 25 steps. Guidance is disabled after the selected prefix.}
\label{fig:component-window-ablation}
\end{figure}

\noindent\textbf{Guidance components.} Both \sfgx{} and \sfgs{} improve generation independently. SFG-X gives larger gains, and their combination performs best (Table~\ref{tab:geneval-nocfg}; Figure~\ref{fig:component-window-ablation}a). Combining them improves GenEval over SFG-X from $0.5621$ to $0.5811$ on SD3M and from $0.6277$ to $0.6853$ on FLUX. T2I-CompBench++ also rises from $0.4505$ to $0.4760$ and from $0.4515$ to $0.4727$, respectively. SFG-S thus adds gains within the same auxiliary evaluation.

\noindent\textbf{Guidance windows.} Longer SFG guidance windows do not consistently improve generation, whereas the tested image baselines improve as their guidance windows grow (Figure~\ref{fig:component-window-ablation}b). FLUX CFG+SFG scores $0.7479$ over steps 1--6 versus $0.7441$ over all 28 steps; SD3.5M CFG+SFG scores $0.7111$ over steps 1--10 versus $0.6941$ over all 40 steps. By comparison, extending guidance from 10 to 28 steps raises FLUX CFG from $0.5675$ to $0.6487$; extending it from 10 to 40 steps raises SD3M SEG from $0.4505$ to $0.5323$, with monotonic gains at the tested intermediate windows. Consistent with the temporal transplantation results, early feedback control matches or exceeds full-schedule SFG while requiring fewer auxiliary evaluations.

The video results show the same pattern for SFG (Figure~\ref{fig:component-window-ablation}c): activating it for the first seven of 25 steps scores $84.25$, compared with $83.09$ over all steps, while requiring 32 rather than 50 model evaluations. CFG changes little across its tested windows ($83.48$--$83.58$), remaining below the early SFG window. The useful SFG window therefore varies by task: the strongest tested T2V setting uses seven of 25 steps, rather than the first 20\% used in the image benchmark comparisons. These results support concentrating feedback control early in sampling; extending it to the full trajectory can reduce quality while adding computation.

\section{Related Work}
\label{sec:related_work}

\noindent\textbf{Negative branches and attention strengthening.} CFG contrasts conditional and unconditional predictions; training-free self-guidance extrapolates away from a perturbed model prediction \citep{ho2022classifierfree,chung2025cfgpp,hong2023sag,ahn2024pag, hong2024seg,chen2026s2guidance,zhang2026ssg,chen2026guidedbridge}. SFG-X constructs this reference by strengthening I2T feedback instead of degrading the model through blurred attention, identity maps, or skipped blocks, using the measured response to select the sign. Limited-interval guidance restricts CFG to an empirically chosen interval \citep{kynkaanniemi2024interval}; our transplantation experiment measures when feedback most affects the outcome. Attend-and-Excite amplifies neglected subject-token activations \citep{chefer2023attendexcite}. TACA uses a timestep-aware temperature to rebalance cross-modal attention and counter text-token imbalance \citep{lv2025taca}. Both strengthen T2I conditioning, as SFG-S does through guidance, but neither modifies how the image writes back into text. SFG-X instead controls the I2T direction, where our intervention sweeps support suppressing feedback.

\noindent\textbf{The two streams of an MMDiT.} Studies of MMDiTs examine padding registers, multimodal circuits, prompt reinjection, and attention editing \citep{toker2025padding,mazur2026diffract,ge2026visionlanguagebinding, zhou20253disflux,shen2025qkedit,yao2026promptreinjection}. \citet{toker2025padding} find image content in padding tokens; \citet{li2026registers} find template tokens maintaining object identity; \citet{shin2025mmdit} identify I2T as the weaker cross-modal direction and build on T2I. We test how the stored content affects generation and find that I2T feedback can preserve errors.

\section{Conclusion}
\label{sec:conclusion}

We show that image-to-text feedback in MMDiTs can preserve image content that conflicts with the prompt. Message transplantation, local response analysis, and matched interventions identify feedback suppression as a useful control direction. Building on these findings, \sfg{} guides away from an auxiliary branch with stronger feedback and weaker text-to-image conditioning (about $1.2\times$ base NFE without CFG), while \isfg{} applies the corresponding controls directly with zero extra NFE. Experiments across three image backbones demonstrate gains in alignment and preference scores, both without and alongside CFG. The same control strategy improves the evaluated VBench scores in text-to-video and image-to-video generation.

\begingroup
\small
\bibliographystyle{iclr2027_conference}
\bibliography{references}
\endgroup

\appendix

\section{Theoretical analysis of path interventions}
\label{app:path-theory}
\begingroup\small
\label{app:guidance-conventions}
\noindent\textbf{Guidance conventions.}

CFG uses condition dropout during training and extrapolates between conditional and unconditional predictions; AG instead contrasts well-trained and under-trained models \citep{ho2022classifierfree,karras2024guiding}. Training-free self-guidance perturbs one pretrained model:
\begin{equation}
 v_{\rm CFG}=v_{\rm uncond}
 +s(v_{\rm clean}-v_{\rm uncond}),
 \label{eq:cfg-background}
\end{equation}
\begin{equation}
 v_{\rm guide}=v_{\rm clean}
 +\omega(v_{\rm clean}-v_{\rm pert}).
 \label{eq:self-guidance-background}
\end{equation}
Here $v_{\rm clean}$ is the clean-branch prediction, $v_{\rm uncond}$ is the unconditional prediction, $s$ is the CFG scale, and $v_{\rm pert}$ supplies the negative branch. SAG blurs image regions, PAG replaces attention with the identity, SEG smooths attention, S$^2$-Guidance drops blocks, and SSG swaps extreme text tokens \citep{hong2023sag,ahn2024pag,hong2024seg,chen2026s2guidance,zhang2026ssg}. We choose the perturbation by measuring a native feedback path.

\noindent\textbf{Local relation between SFG and iSFG.} Let $\mathbf a=(a_X,a_S)$ denote the I2T and T2I message multipliers: $(1-u,1-|u|)$ in the SFG auxiliary branch, $(1+\gamma,1+\gamma_{\rm S})$ in the iSFG conditional branch, and $(\alpha,1)$ in the local-response experiment. At fixed latent, time, and prompt, write the prediction as $v(\mathbf a)$. The clean branch uses $\mathbf a_0=(1,1)$, and the auxiliary branch uses $\mathbf a_0+\mathbf h$, where $\mathbf h=(d_X,-d_S)$ and $d_X,d_S\geq0$. For a smooth prediction, the first-order correction is
\begin{equation}
\omega[v(\mathbf a_0)-v(\mathbf a_0+\mathbf h)]
=-\omega d_XJ_X+\omega d_SJ_S+O(\omega\|\mathbf h\|^2),
\label{eq:theory-tangent-bound}
\end{equation}
where $J_X$ and $J_S$ are the derivatives with respect to the two message multipliers at $\mathbf a_0$. SFG-X uses $d_S=0$, and full SFG uses $d_X=d_S=|u|$. Setting the message multipliers to $\mathbf a_0-\omega\mathbf h$ gives the same first-order correction in place. With CFG scale $s>0$ and an unchanged unconditional branch, the matching conditional-branch multipliers are $\mathbf a_0-(\omega/s)\mathbf h$. This local relation motivates iSFG; the separately tuned methods need not agree at finite strengths.

\noindent\textbf{Connection to the measured response.} The image-to-text derivative $J_X$ corresponds to $J$ in the alignment experiment. Appendix~\ref{app:jacobian} gives its centered-difference estimate and the successful-donor velocity response $R^S$. Reducing feedback moves locally along $-J$. A positive cosine gap over random indicates better relative alignment; it does not establish positive absolute cosine or improved final quality. Final outcomes and intervention windows are evaluated by transplantation and the sign ablation, rather than inferred from the local expansion. Probe definitions and their aggregation are given with the probe results in App.~\ref{app:object-probes}.

\begingroup
\setlength{\jot}{2pt}
\begin{equation}
\begin{aligned}
J_t&=\left.\partial_\alpha v(x_t^F,t,c;\alpha)\right|_{\alpha=1},\\
R_t^S&=v(x_t^F,t,c;B_t\leftarrow B_t^S)-v(x_t^F,t,c;B_t\leftarrow B_t^F),\\
\cos(-J_t,R_t^S)&=\frac{\langle-J_t,R_t^S\rangle}{\|J_t\|_2\,\|R_t^S\|_2},\\
\Delta\cos_t&=\cos(-J_t,R_t^S)-\cos(-J_t,R_t^{\rm rand}).
\end{aligned}
\label{eq:jacobian-rescue}
\end{equation}
\endgroup

\noindent\textbf{Control-direction measurements.} Across 256 prompts and 40 sampling steps, SFG-X and SFG-S have low mutual cosine similarity; their similarities with CFG also remain well below one (Figure~\ref{fig:orthogonality}). These distinct correction directions are consistent with the gains from combining the controls.
\begin{figure}[H]
\centering
\includegraphics[width=0.95\linewidth]{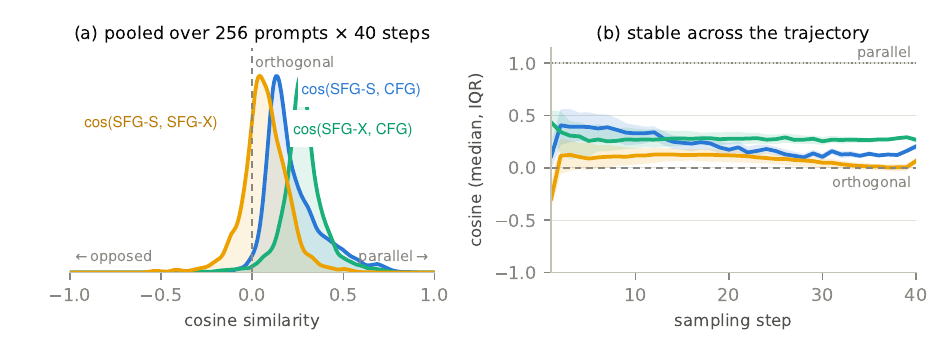}
\caption{\textbf{Complementary directions.} \sfgx{} and \sfgs{} corrections have low cosine similarity with CFG and with each other, both over prompts and through sampling.}
\label{fig:orthogonality}
\end{figure}

\endgroup
 
\section{Benchmark evaluation and parameter settings}
\label{app:hparams}
Within each model and evaluation configuration, all methods share the prompts, seed set, and initial noise. Table~\ref{tab:benchmark-sample-counts} lists distinct prompt counts and generation seeds separately. Each prompt--seed pair produces one sample; for I2V, each prompt also has a fixed reference image. ImageReward scores the same generated images as HPSv2.1.
\begin{table}[H]
\centering\small
\setlength{\tabcolsep}{4pt}
\caption{\textbf{Evaluation sample counts per model and method configuration.} Seeds/prompt gives the number of generation seeds for each distinct prompt, or prompt--reference-image pair for I2V. The 800 HPSv2.1 prompts are the total across four styles; four seeds per prompt yield 3,200 images. ImageReward reuses these images without additional generation.}
\label{tab:benchmark-sample-counts}
\begin{tabular}{llrr}\toprule
Benchmark & Prompts & Seeds/prompt & Total outputs\\\midrule
GenEval & $553$ & $6$ & $3{,}318$\\
T2I-CompBench++ & $300$ & $8$ & $2{,}400$\\
HPSv2.1 & $800$ & $4$ & $3{,}200$\\
ImageReward & $800$ & $4$ & $3{,}200$\\
VBench (T2V) & $326$ & $3$ & $978$\\
VBench (I2V) & $355$ & $3$ & $1{,}065$\\
\bottomrule\end{tabular}
\end{table}
\noindent\textbf{Generation settings.} Images use $1024^2$ resolution. Image benchmarks and guidance-window ablations use 28 steps for FLUX and 40 for SD3M/SD3.5M; the transplantation experiments in App.~\ref{app:temporal} use 40 steps for both models. HunyuanVideo T2V/I2V use $848\times480$, 121 frames, 24 fps, 25 steps, Flow shift 5.0, and BF16; I2V uses matched $16{:}9$ reference crops. Their $(|u|,\omega)$ settings are $(0.25,4)$ and $(0.35,6)$, respectively. Explicit \sfg{}/\sfgx{} acts in all joint-attention layers over a guidance window. Image benchmark comparisons use the first $20\%$ of steps; the video window is set for the task, with seven of 25 steps used for the main T2V result. \isfg{} acts at every step.

\makeatletter
\newcommand{\ownscaptype}[1]{\def\@captype{#1}}
\makeatother
\noindent\textbf{SFG guidance-scale sweeps.} We report recorded evaluations across guidance scales $\omega$. Curves connect the recorded points in order of guidance scale. Bold entries and plot annotations mark the highest recorded score in each table. Each point uses the seed set specified in Table~\ref{tab:benchmark-sample-counts}; GenEval uses six seeds per prompt.
\par\medskip\noindent\begin{minipage}[c]{0.37\linewidth}\centering\small
\ownscaptype{table}
\caption{\textbf{SFG scale sweep on SD3M without CFG} \citep{ho2022classifierfree}. GenEval versus guidance scale $\omega$; bold marks the best tested score.}
\label{tab:own-sweep-sd3_nocfg}
\label{tab:sfg-recorded-scales}
\begin{tabular}{cc}\toprule
$\omega$ & GenEval $\uparrow$\\\midrule
\textbf{5} & \textbf{0.5811}\\
6 & 0.5787\\
7 & 0.5751\\
8 & 0.5730\\
\bottomrule\end{tabular}
\end{minipage}\hfill
\begin{minipage}[c]{0.60\linewidth}\centering
\includegraphics[width=\linewidth]{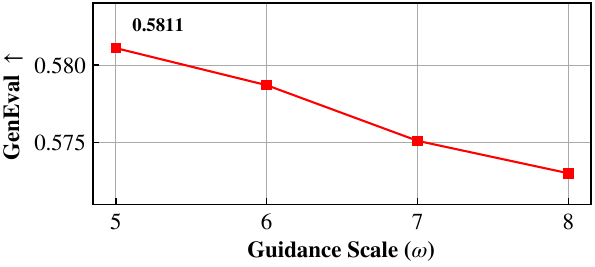}
\ownscaptype{figure}
\caption{SD3M without CFG / GenEval guidance-scale sweep.}
\label{fig:own-sweep-sd3_nocfg}
\end{minipage}\par\vspace{0.5em}
\par\medskip\noindent\begin{minipage}[c]{0.37\linewidth}\centering\small
\ownscaptype{table}
\caption{\textbf{SFG scale sweep on FLUX without CFG} \citep{ho2022classifierfree}. GenEval versus $\omega$, with built-in guidance set to 1; bold marks the best tested score.}
\label{tab:own-sweep-flux_nocfg}
\begin{tabular}{cc}\toprule
$\omega$ & GenEval $\uparrow$\\\midrule
7 & 0.6065\\
9 & 0.6233\\
10 & 0.6475\\
11 & 0.6655\\
12 & 0.6739\\
13 & 0.6779\\
\textbf{14} & \textbf{0.6853}\\
\bottomrule\end{tabular}
\end{minipage}\hfill
\begin{minipage}[c]{0.60\linewidth}\centering
\includegraphics[width=\linewidth]{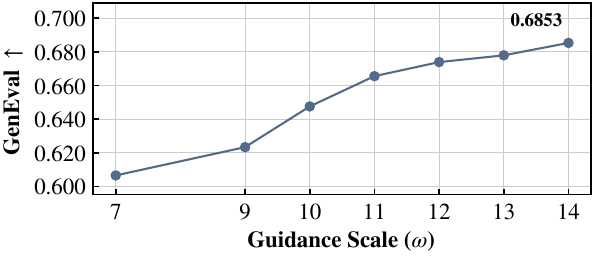}
\ownscaptype{figure}
\caption{FLUX without CFG / GenEval guidance-scale sweep.}
\label{fig:own-sweep-flux_nocfg}
\end{minipage}\par\vspace{0.5em}
\noindent The non-CFG records use FLUX.1-dev with built-in guidance 1.
\noindent\textbf{FLUX stacked on CFG.} We report GenEval and HPSv2.1 separately. The GenEval records use de-distilled FLUX with explicit CFG scale 3.5.
\par\medskip\noindent\begin{minipage}[c]{0.37\linewidth}\centering\small
\ownscaptype{table}
\caption{\textbf{SFG scale sweep on FLUX with CFG} \citep{ho2022classifierfree}. GenEval versus $\omega$ at CFG scale 3.5; bold marks the best tested score.}
\label{tab:own-sweep-flux_cfg_geneval}
\begin{tabular}{cc}\toprule
$\omega$ & GenEval $\uparrow$\\\midrule
7 & 0.7421\\
\textbf{7.5} & \textbf{0.7479}\\
8 & 0.7432\\
8.5 & 0.7442\\
\bottomrule\end{tabular}
\end{minipage}\hfill
\begin{minipage}[c]{0.60\linewidth}\centering
\includegraphics[width=\linewidth]{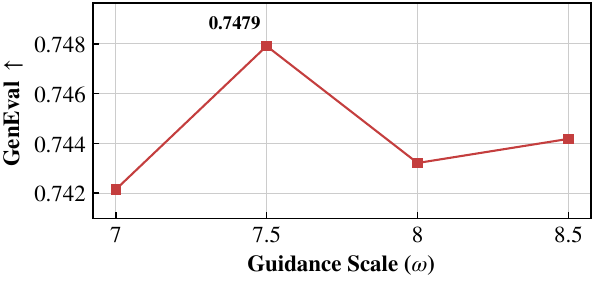}
\ownscaptype{figure}
\caption{FLUX with CFG / GenEval guidance-scale sweep.}
\label{fig:own-sweep-flux_cfg_geneval}
\end{minipage}\par\vspace{0.5em}
\par\medskip\noindent\begin{minipage}[c]{0.37\linewidth}\centering\small
\ownscaptype{table}
\caption{\textbf{SFG scale sweep on FLUX with CFG} \citep{ho2022classifierfree}. HPSv2.1 versus $\omega$; bold marks the best tested score.}
\label{tab:own-sweep-flux_cfg_hps}
\begin{tabular}{cc}\toprule
$\omega$ & HPSv2.1 $\uparrow$\\\midrule
5 & 31.54\\
5.5 & 31.54\\
6 & 31.58\\
6.5 & 31.57\\
\textbf{7.5} & \textbf{31.66}\\
\bottomrule\end{tabular}
\end{minipage}\hfill
\begin{minipage}[c]{0.60\linewidth}\centering
\includegraphics[width=\linewidth]{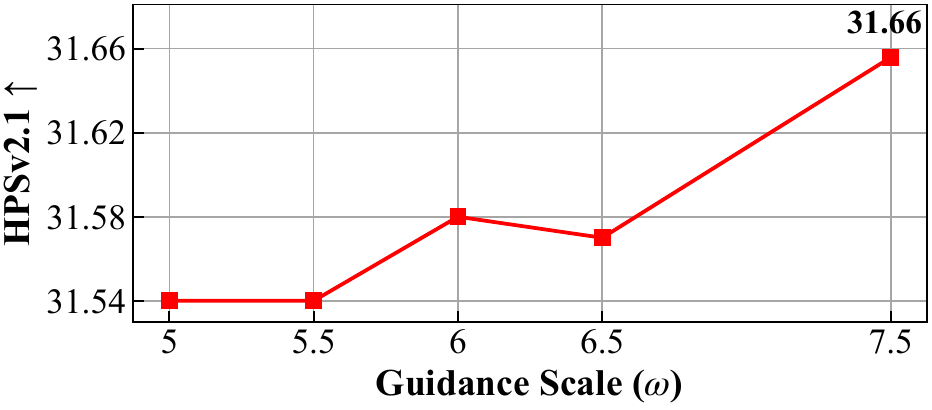}
\ownscaptype{figure}
\caption{FLUX with CFG / HPSv2.1 guidance-scale sweep.}
\label{fig:own-sweep-flux_cfg_hps}
\end{minipage}\par\vspace{0.5em}
 
\begingroup
\raggedbottom
\section{Baseline hyperparameter sweeps}
\label{app:baseline-sweeps}

Each GenEval sweep point uses 553 prompts with six seeds per prompt (3,318 images), matching App.~\ref{app:hparams}, at $1024\times1024$ resolution with the official Mask2Former evaluator and 40 steps for SD3M/SD3.5M or 28 for FLUX. \textbf{Negative-branch operators:} PAG replaces selected attention maps by identity ($A\leftarrow I$), removing cross-token mixing \citep{ahn2024pag}; SEG Gaussian-blurs spatial queries ($Q\leftarrow G_\sigma*Q$), smoothing attention logits before softmax \citep{hong2024seg}; S$^2$ stochastically skips Transformer blocks \citep{chen2026s2guidance}. Our sweeps fix PAG at block 8, SEG at blocks 7/8/9 with $\sigma=9{,}999{,}999$, and S$^2$ to the all-block range. Only $w$ varies within each curve. Five-point displays retain the tested endpoints and observed maxima; SD3.5M PAG/SEG instead show six/three points. Bold entries mark observed maxima; source records accompany the manuscript. The updated main-table GenEval rows use the best confirmed sweep points.

\makeatletter
\newcommand{\sweepcaptype}[1]{\def\@captype{#1}}
\makeatother
\subsection{SD3M without CFG}
\label{app:sweeps-sd3}
The CFG scale is 1.0 and the matched GenEval reference is 0.3294. S$^2$ uses drop 1 during steps 5--36. All three methods improve on the reference; PAG reaches 0.5604 at $w=3$ among the completed fixed-configuration points.
\par\vspace{0.5em}
\noindent\begin{minipage}[c]{0.37\linewidth}
\centering\small
\sweepcaptype{table}
\caption{\textbf{PAG on SD3M without CFG.} \citep{ahn2024pag} GenEval versus guidance weight $w$; bold marks the best tested setting ($w=3$).}
\label{tab:sweep-sd3-pag}\label{tab:baseline-sweeps}
\begin{tabular}{cc}\toprule
Scale ($w$) & GenEval $\uparrow$\\\midrule
0.5 & 0.4696\\
1.25 & 0.5521\\
2.5 & 0.5580\\
\textbf{3} & \textbf{0.5604}\\
4 & 0.5563\\
\bottomrule\end{tabular}
\end{minipage}\hfill
\begin{minipage}[c]{0.60\linewidth}
\centering
\includegraphics[width=\linewidth]{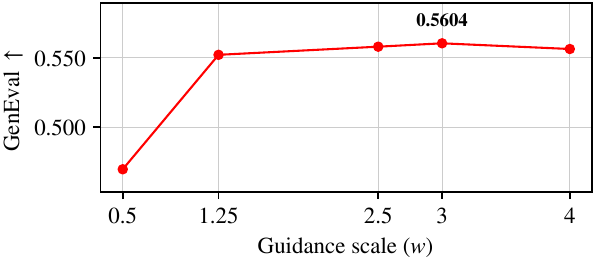}
\sweepcaptype{figure}
\caption{\textbf{PAG.} Five-point weight sweep.}
\label{fig:sweep-sd3-pag}\label{fig:baseline-sweeps}
\end{minipage}\par\vspace{0.8em}

\noindent\begin{minipage}[c]{0.37\linewidth}
\centering\small
\sweepcaptype{table}
\caption{\textbf{SEG on SD3M without CFG.} \citep{hong2024seg} GenEval versus guidance weight $w$; bold marks the best tested setting ($w=1$).}
\label{tab:sweep-sd3-seg}
\begin{tabular}{cc}\toprule
Scale ($w$) & GenEval $\uparrow$\\\midrule
0.2 & 0.4264\\
0.5 & 0.4684\\
\textbf{1} & \textbf{0.5322}\\
1.25 & 0.5303\\
3 & 0.5212\\
\bottomrule\end{tabular}
\end{minipage}\hfill
\begin{minipage}[c]{0.60\linewidth}
\centering
\includegraphics[width=\linewidth]{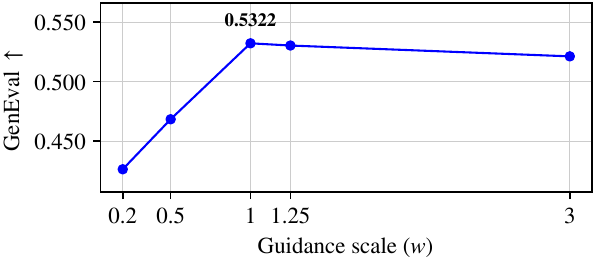}
\sweepcaptype{figure}
\caption{\textbf{SEG.} Five-point weight sweep.}
\label{fig:sweep-sd3-seg}
\end{minipage}\par\vspace{0.8em}

\noindent\begin{minipage}[c]{0.37\linewidth}
\centering\small
\sweepcaptype{table}
\caption{\textbf{S$^2$-Guidance on SD3M without CFG.} \citep{chen2026s2guidance} GenEval versus guidance weight $w$; bold marks the best tested setting ($w=2$).}
\label{tab:sweep-sd3-s2}
\begin{tabular}{cc}\toprule
Scale ($w$) & GenEval $\uparrow$\\\midrule
0.2 & 0.3814\\
0.5 & 0.4415\\
1 & 0.4646\\
\textbf{2} & \textbf{0.4866}\\
3 & 0.4754\\
\bottomrule\end{tabular}
\end{minipage}\hfill
\begin{minipage}[c]{0.60\linewidth}
\centering
\includegraphics[width=\linewidth]{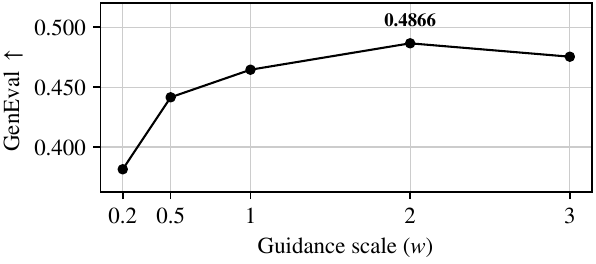}
\sweepcaptype{figure}
\caption{\textbf{S$^2$.} Five-point weight sweep.}
\label{fig:sweep-sd3-s2}
\end{minipage}\par\vspace{0.8em}

\subsection{SD3.5M with CFG}
\label{app:sweeps-sd35}
The CFG scale is 7.5 and the matched GenEval reference is 0.6890. S$^2$ uses drop 3 during steps 5--36. PAG peaks at 0.6807 for $w=0.1$, and the displayed S$^2$ maximum is 0.6811; both remain below the reference. SEG reaches 0.6890 at $w=0.2$, close to the reference.
\par\vspace{0.5em}
\noindent\begin{minipage}[c]{0.37\linewidth}
\centering\small
\sweepcaptype{table}
\caption{\textbf{PAG on SD3.5M with CFG.} \citep{ahn2024pag} GenEval versus guidance weight $w$; bold marks the best tested setting ($w=0.1$).}
\label{tab:sweep-sd35-pag}
\begin{tabular}{cc}\toprule
Scale ($w$) & GenEval $\uparrow$\\\midrule
0.05 & 0.6729\\
\textbf{0.1} & \textbf{0.6807}\\
0.2 & 0.6669\\
0.75 & 0.6650\\
1 & 0.6568\\
1.5 & 0.6499\\
\bottomrule\end{tabular}
\end{minipage}\hfill
\begin{minipage}[c]{0.60\linewidth}
\centering
\includegraphics[width=\linewidth]{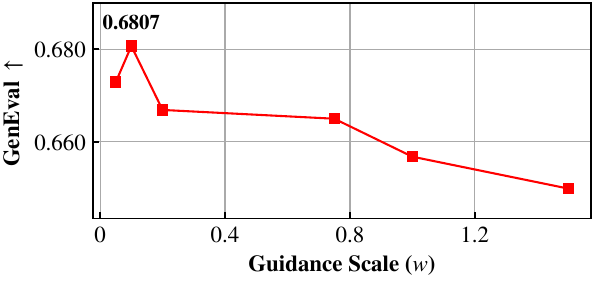}
\sweepcaptype{figure}
\caption{\textbf{PAG.} Six-point weight sweep.}
\label{fig:sweep-sd35-pag}
\end{minipage}\par\vspace{0.8em}

\noindent\begin{minipage}[c]{0.37\linewidth}
\centering\small
\sweepcaptype{table}
\caption{\textbf{SEG on SD3.5M with CFG.} \citep{hong2024seg} GenEval versus guidance weight $w$; bold marks the best tested setting ($w=0.2$).}
\label{tab:sweep-sd35-seg}
\begin{tabular}{cc}\toprule
Scale ($w$) & GenEval $\uparrow$\\\midrule
0.05 & 0.6729\\
\textbf{0.2} & \textbf{0.6890}\\
0.5 & 0.6741\\
\bottomrule\end{tabular}
\end{minipage}\hfill
\begin{minipage}[c]{0.60\linewidth}
\centering
\includegraphics[width=\linewidth]{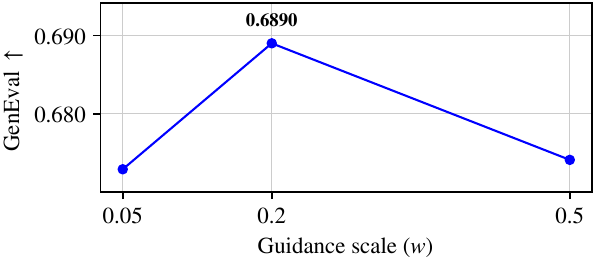}
\sweepcaptype{figure}
\caption{\textbf{SEG.} Three-point weight sweep.}
\label{fig:sweep-sd35-seg}
\end{minipage}\par\vspace{0.8em}

\noindent\begin{minipage}[c]{0.37\linewidth}
\centering\small
\sweepcaptype{table}
\caption{\textbf{S$^2$-Guidance on SD3.5M with CFG.} \citep{chen2026s2guidance} GenEval versus guidance weight $w$; bold marks the best tested setting ($w=0.25$).}
\label{tab:sweep-sd35-s2}
\begin{tabular}{cc}\toprule
Scale ($w$) & GenEval $\uparrow$\\\midrule
\textbf{0.25} & \textbf{0.6811}\\
1.5 & 0.6274\\
2 & 0.5846\\
3 & 0.4977\\
4.5 & 0.4227\\
\bottomrule\end{tabular}
\end{minipage}\hfill
\begin{minipage}[c]{0.60\linewidth}
\centering
\includegraphics[width=\linewidth]{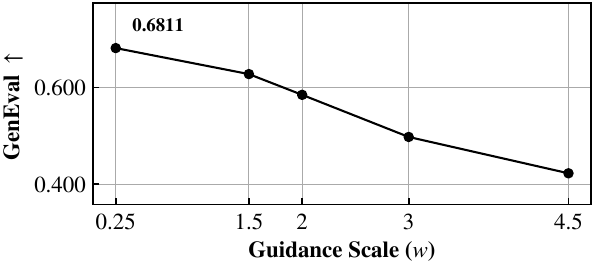}
\sweepcaptype{figure}
\caption{\textbf{S$^2$.} Five-point weight sweep.}
\label{fig:sweep-sd35-s2}
\end{minipage}\par\vspace{0.8em}

\subsection{Original FLUX.1-dev (g = 1.0)}
\label{app:sweeps-flux}
The original FLUX.1-dev model uses $g=1.0$, with a matched GenEval reference of 0.3604. S$^2$ uses drop 5 during steps 4--25. All three methods improve on the reference; SEG reaches 0.5548 at $w=3$. These points use the original model and are separate from de-distilled FLUX.
\par\vspace{0.5em}
\noindent\begin{minipage}[c]{0.37\linewidth}
\centering\small
\sweepcaptype{table}
\caption{\textbf{PAG on FLUX without CFG.} \citep{ahn2024pag} GenEval versus guidance weight $w$; bold marks the best tested setting ($w=2$).}
\label{tab:sweep-flux-pag}
\begin{tabular}{cc}\toprule
Scale ($w$) & GenEval $\uparrow$\\\midrule
1 & 0.4626\\
1.5 & 0.4728\\
\textbf{2} & \textbf{0.4866}\\
2.5 & 0.4706\\
3 & 0.4757\\
\bottomrule\end{tabular}
\end{minipage}\hfill
\begin{minipage}[c]{0.60\linewidth}
\centering
\includegraphics[width=\linewidth]{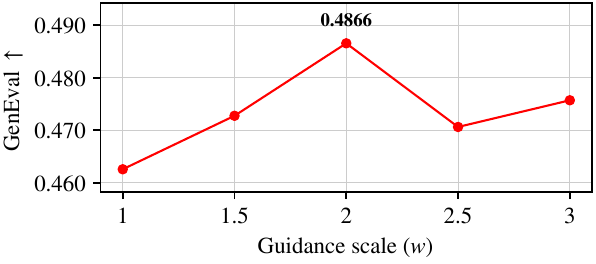}
\sweepcaptype{figure}
\caption{\textbf{PAG.} Five-point weight sweep.}
\label{fig:sweep-flux-pag}
\end{minipage}\par\vspace{0.8em}

\noindent\begin{minipage}[c]{0.37\linewidth}
\centering\small
\sweepcaptype{table}
\caption{\textbf{SEG on FLUX without CFG.} \citep{hong2024seg} GenEval versus guidance weight $w$; bold marks the best tested setting ($w=3$).}
\label{tab:sweep-flux-seg}
\begin{tabular}{cc}\toprule
Scale ($w$) & GenEval $\uparrow$\\\midrule
2 & 0.5315\\
2.5 & 0.5295\\
\textbf{3} & \textbf{0.5548}\\
3.5 & 0.5544\\
4 & 0.5525\\
\bottomrule\end{tabular}
\end{minipage}\hfill
\begin{minipage}[c]{0.60\linewidth}
\centering
\includegraphics[width=\linewidth]{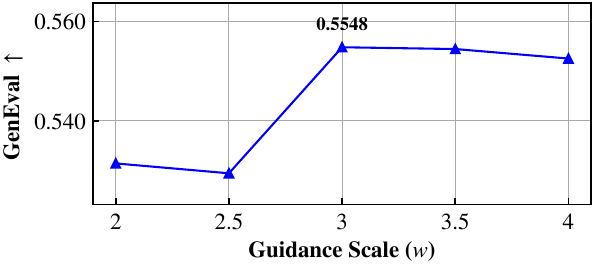}
\sweepcaptype{figure}
\caption{\textbf{SEG.} Five-point weight sweep.}
\label{fig:sweep-flux-seg}
\end{minipage}\par\vspace{0.8em}

\noindent\begin{minipage}[c]{0.37\linewidth}
\centering\small
\sweepcaptype{table}
\caption{\textbf{S$^2$-Guidance on FLUX without CFG.} \citep{chen2026s2guidance} GenEval versus guidance weight $w$; bold marks the best tested setting ($w=2$).}
\label{tab:sweep-flux-s2}
\begin{tabular}{cc}\toprule
Scale ($w$) & GenEval $\uparrow$\\\midrule
1 & 0.4876\\
\textbf{2} & \textbf{0.4990}\\
2.5 & 0.4709\\
3 & 0.4548\\
3.5 & 0.4347\\
\bottomrule\end{tabular}
\end{minipage}\hfill
\begin{minipage}[c]{0.60\linewidth}
\centering
\includegraphics[width=\linewidth]{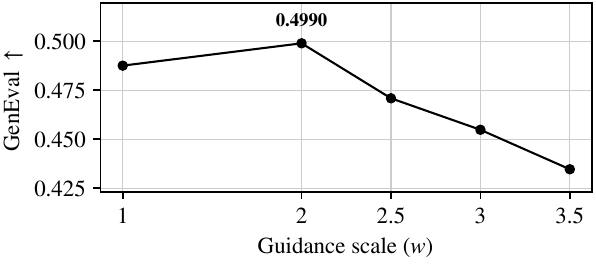}
\sweepcaptype{figure}
\caption{\textbf{S$^2$.} Five-point weight sweep.}
\label{fig:sweep-flux-s2}
\end{minipage}\par\vspace{0.8em}

\endgroup

\section{Implementation details and pseudocode}
\label{app:impl}

All experiments use one attention-processor patch on the public \texttt{diffusers} implementations of SD3-Medium, SD3.5-Medium, and FLUX. The patch touches nothing but the fused scaled-dot-product attention (SDPA) call of each joint-attention block:

\begin{small}
\begin{verbatim}
# One MMDiT joint-attention block. X: image tokens, C: text tokens.
Qx, Kx, Vx = qkv_image(X)               # modality-specific projections
Qc, Kc, Vc = qkv_text(C)
Q, K, V    = concat(Qx,Qc), concat(Kx,Kc), concat(Vx,Vc)
O          = SDPA(Q, K, V)              # one joint softmax

# Cross-modal term of a chosen direction, at the SAME softmax weights:
# zero every value row outside the source modality, rerun SDPA on the
# target-query slice only.
B_i2t = SDPA(Q[text],  K, V * mask(image))  # = A_{i->t} V_i
B_t2i = SDPA(Q[image], K, V * mask(text))   # = A_{t->i} V_t

O[text]  += (a_X - 1) * B_i2t  # I2T message multiplier a_X
O[image] += (a_S - 1) * B_t2i  # T2I message multiplier a_S
# ... followed by the block's usual output projections.
\end{verbatim}
\end{small}

Because the mask only zeroes value rows, the softmax weights are exactly those of the unmodified forward, and the added term is exactly $A_{\mathrm{source}\rightarrow\mathrm{target}}V_{\mathrm{source}}$. Adding $(a_X-1)B_{\img\rightarrow\txt}$ or $(a_S-1)B_{\txt\rightarrow\img}$ sets the corresponding message multiplier to $a_X$ or $a_S$. The sampler-level combination is then

\begin{small}
\begin{verbatim}
# Explicit SFG: use the guidance window selected for the task.
# K_act: e.g. 8/40 image steps or 7/25 T2V steps.
if step < K_act:
    v_clean = v_theta(x_t, t, c)
    # Full SFG (u < 0); SFG-X alone sets a_S = 1
    v_pert  = v_theta(x_t, t, c, a_X=1-u, a_S=1-abs(u))
    v_hat   = v_clean + omega * (v_clean - v_pert)  # + CFG if stacked
else:
    v_hat   = v_theta(x_t, t, c)       # ordinary forward, no extra NFE
# iSFG: gamma < 0 for I2T, gamma_S > 0 for T2I
v_ip = v_theta(x_t, t, c, a_X=1+gamma, a_S=1+gamma_S)
# Use v_ip directly, or combine with the unedited unconditional
# prediction through CFG; no auxiliary branch is evaluated.
\end{verbatim}
\end{small}

The path edit is applied in every joint-attention block (\texttt{layers=all}). For explicit \sfg{}, the negative branch is evaluated only during the selected guidance window, batched with the clean branch in a single fused forward; the remaining steps run the ordinary forward with no extra evaluation. The separate probe diagnostics in App.~\ref{app:object-probes} use their explicitly specified windows, including full-schedule guidance. The in-place \isfg{} edit runs at every step at zero extra NFE. A norm-clipping safeguard on the residual ($\mathrm{clip}=\min(1,\tau\|v_{\boldsymbol\theta}(x_t,t,c;\mathbf 1)\|/\|\Delta\|)$) exists in the implementation but is disabled ($\tau{=}999$) in the benchmark comparisons.

\par\medskip
\noindent\begin{minipage}{\linewidth}
\small
\subsection{Low-overhead iSFG implementation}
\label{app:isfg-runtime}
The optimized implementation folds the message multipliers into value aggregation, without a separate auxiliary Transformer branch or recomputation of the cross-modal messages above. Image queries use the original image values and text values scaled by $1.20$; text queries use the original text values and image values scaled by $0.80$. Both query groups attend to all keys with the original joint-softmax normalization, and their outputs are concatenated in token order. The disjoint query groups cover the original joint attention without duplicating its attention-score computation. Under CFG, only the conditional branch is edited. Remaining overhead comes from value copies, scaling, and kernel dispatch.

We time SD3.5M on one A800 80GB GPU with BF16, $1024^2$ resolution, 40 steps, CFG scale 7.5, and batch size 1. iSFG uses $(\gamma_{\rm S},\gamma)=(+0.20,-0.20)$ in all joint blocks at all steps. After one warm-up per method, we alternate three paired runs with matched prompts and seeds. Timing includes text encoding, denoising, and VAE decoding, excluding model loading and image writes. Both methods make 40 batched conditional/unconditional Transformer calls. iSFG adds about $0.31$ seconds per image ($2.86\%$), with zero extra NFE.

\begin{table}[H]
\centering\small
\caption{\textbf{iSFG runtime.} Mean $\pm$ standard deviation over three paired runs.}
\label{tab:isfg-runtime}
\begin{tabular}{lccc}
\toprule
Method & Extra NFE & Time per image (s) & Relative time \\
\midrule
CFG & 0 & $10.800\pm0.020$ & $1.000\times$ \\
CFG+iSFG (ours) & 0 & $11.108\pm0.018$ & $1.029\times$ \\
\bottomrule
\end{tabular}
\end{table}

\end{minipage}\par\medskip

\section{Paired sampling protocol}
\label{app:stats}
For each benchmark and backbone, all baselines and our methods use the same prompts and seed set. Each prompt is generated with multiple seeds before scoring; every generated sample is evaluated, and scores are aggregated according to the benchmark protocol. GenEval uses 553 prompts with six seeds per prompt (3,318 images per configuration); the counts for other benchmarks are listed in Table~\ref{tab:benchmark-sample-counts}. Methods share the initial noise for each prompt--seed pair, and I2V also fixes the reference image. For each method on a given backbone, we use the same hyperparameters across all applicable benchmarks, with no benchmark-specific tuning. Every hyperparameter-sweep point is evaluated on the full test sets of all applicable benchmarks using the same seed set. Mechanism diagnostics retain their separately stated sampling settings.

\noindent\textbf{Seed variability.} The reported standard deviations across seeds are $0.031$ for VBench and $0.00332$ for GenEval, in the same units as their reported overall scores.

\section{Full-model intervention and sign checks}
\label{app:sweep}
\subsection{I2T message and temporal transplantation}
\label{app:b-definition}
\label{app:temporal}
The transplantation experiment uses 400 same-prompt success/failure pairs across SD3.5M and FLUX, at $1024^2$, $\mathrm{cfg}=1$, and 40 steps, editing all joint-attention blocks. Main-text success criteria are rechecked during replay; conditions share the recipient prompt, initial latent, and sampler. Let $k,t_k,\ell$ denote sampling step, scheduler timestep, and block, and $d_h$ the head dimension. With image tokens preceding text tokens,
\begin{equation}
 K=[K_{\img};K_{\txt}],\qquad V=[V_{\img};V_{\txt}],\qquad
 A_{\txt}=\operatorname{softmax}\!\left(Q_{\txt}K^\top/\sqrt{d_h}\right).
 \label{eq:app-b-attention}
\end{equation}
\begin{equation}
 B_{\img\rightarrow\txt}=A_{\txt}[V_{\img};0],\qquad
 O_{\txt}=B_{\img\rightarrow\txt}+A_{\txt}[0;V_{\txt}].
 \label{eq:app-b-decomposition}
\end{equation}
\begin{equation}
 \widetilde O_{\txt}=O_{\txt}-B+\widetilde B.
 \label{eq:app-b-replace}
\end{equation}
$B$ retains the full joint-softmax denominator. Only its image-to-text contribution is replaced after value aggregation and before text output projection. Cache $B^S_{k,\ell}$ on the native successful rollout and align by step/block; do not recompute it with recipient queries. For enabled steps/blocks $\mathcal W,\mathcal L$ and current recipient message $B^{\mathcal W}_{k,\ell}$,
\begin{equation}
 \widetilde B^{\mathcal W}_{k,\ell}
 =B^{\mathcal W}_{k,\ell}
 +\lambda\,\mathbf 1\{k\in\mathcal W,\ell\in\mathcal L\}
       (B^S_{k,\ell}-B^{\mathcal W}_{k,\ell}).
 \label{eq:app-window-patch}
\end{equation}
\begin{equation}
 x^{\mathcal W}_{k+1}=\operatorname{Step}
 \bigl(x^{\mathcal W}_k,t_k,v(x^{\mathcal W}_k,t_k,c;\widetilde B^{\mathcal W}_k)\bigr).
 \label{eq:app-patch-scheduler}
\end{equation}
$\lambda=1$ fully transplants and $\lambda=0$ gives the baseline. Restart each condition from the same failed initial state; outside the window use the current recipient message. Steps run from noise (1) to image (40). Full, window, prefix, and suffix sets are $[1,K]$, $[a,b]$, $[1,b]$, and $[a,K]$; Figure~\ref{fig:temporal} uses eight-step windows. With benchmark success criterion $E_i$, evaluate each window separately:
\begin{equation}
 \operatorname{Rescue}(\mathcal W)=
 \frac{\sum_i\mathbf 1\{E_i(I_i^F)=0,\ E_i(I_i^{\mathcal W})=1\}}
      {\sum_i\mathbf 1\{E_i(I_i^F)=0\}}.
 \label{eq:app-rescue-rate}
\end{equation}
\noindent\textbf{Donor and path controls.} For the failed-donor control, we form 200 failure/failure pairs and transplant $B_{\img\rightarrow\txt}$ from one failed trajectory into the other. This rescues $38/200$ recipients (19.0\%) over steps 1--40. For the path control, we retain the same 400 success/failure pairs and successful-donor/failed-recipient assignments, replacing the text-to-image message:
\begin{equation}
 A_{\img}=\operatorname{softmax}\!\left(Q_{\img}K^\top/\sqrt{d_h}\right),\qquad
 B_{\txt\rightarrow\img}=A_{\img}[0;V_{\txt}].
\end{equation}
For this text-to-image (image-read-text) control, full replacement ($\lambda=1$) over steps 1--40 rescues $74/400$ recipients (18.5\%). Over steps 1--8, rescue is $151/400$ (37.8\%) at $\lambda=1$ and $118/400$ (29.5\%) at $\lambda=0.5$. Table~\ref{tab:transplant-controls} reports the window and outcome for each condition.

\begin{table}[htbp]
\centering\small
\setlength{\tabcolsep}{3pt}
\caption{\textbf{Message transplantation and controls on paired generation trajectories.} $S$ and $F$ denote successful and failed trajectories; arrows run from donor to recipient. Image-to-text and text-to-image transplantation use the same 400 success/failure pairs, while failed-to-failed transplantation uses 200 failure/failure pairs. $\lambda$ is the message mixing weight. Steps specify the transplant window. Counts and percentages report rescued failures or disrupted successes.}
\label{tab:transplant-controls}
\begin{tabular*}{\linewidth}{@{\extracolsep{\fill}}lllrl@{}}
\toprule
Message path & Donor $\to$ recipient & Steps & Outcome & Count (rate)\\
\midrule
Image-to-text & $S\to F$, $\lambda=1$ & 1--40 & Rescue & 270/400 (67.5\%)\\
Image-to-text & $S\to F$, $\lambda=0.5$ & 1--40 & Rescue & 207/400 (51.8\%)\\
Image-to-text & Random, $\lambda=0.5$ & 1--40 & Rescue & 5/400 (1.3\%)\\
Image-to-text & $F\to F$ & 1--40 & Rescue & 38/200 (19.0\%)\\
Text-to-image & $S\to F$, $\lambda=1$ & 1--40 & Rescue & 74/400 (18.5\%)\\
Text-to-image & $S\to F$, $\lambda=1$ & 1--8 & Rescue & 151/400 (37.8\%)\\
Text-to-image & $S\to F$, $\lambda=0.5$ & 1--8 & Rescue & 118/400 (29.5\%)\\
\midrule
Image-to-text & $F\to S$, $\lambda=1$ & 1--40 & Disruption & 242/400 (60.5\%)\\
Image-to-text & $F\to S$, $\lambda=0.5$ & 1--40 & Disruption & 232/400 (58.0\%)\\
\bottomrule
\end{tabular*}
\end{table}

\subsection{Local response alignment and aggregation}
\label{app:jacobian}
\label{app:intervention-statistics}
All response measurements share the native failed latent $x_k^F$, timestep, embeddings, and weights. Each block scales its current message as $O_{\txt}(\alpha)=O_{\txt}-B+\alpha B$. $J$ differentiates the full prediction with respect to I2T multiplier $\alpha$; its clean-branch value is one. The finite-difference step size is $\delta=0.05$ in Figure~\ref{fig:temporal}:
\begin{equation}
 \begin{aligned}
 v_k^0&=v(x_k^F,t_k,c;1),&
 v_k^{\pm}&=v(x_k^F,t_k,c;1\pm\delta),\\
 J_k^{(\delta)}&=\frac{v_k^+-v_k^-}{2\delta},&
 J_k&=\left.\partial_\alpha v(x_k^F,t_k,c;\alpha)\right|_{\alpha=1}.
 \end{aligned}
 \label{eq:app-centered-j}
\end{equation}
\begin{equation}
 x^F_{k+1}=\operatorname{Step}(x^F_k,t_k,v_k^0).
 \label{eq:app-native-scheduler}
\end{equation}
Here $v_k^0$ is the clean-branch prediction at the failed latent. Only this prediction advances the original trajectory; the other forwards measure local responses. Anchor the fixed-seed random control to $B^F$ recorded in a separate baseline forward and match message-space norms per step/block:
\begin{equation}
 \begin{aligned}
 D_{k,\ell}&=B^S_{k,\ell}-B^F_{k,\ell},\qquad
 U_{k,\ell}\sim\mathcal N(0,I),\\
 \xi_{k,\ell}&=U_{k,\ell}
 \frac{\|D_{k,\ell}\|_F}{\max(\|U_{k,\ell}\|_F,\epsilon)},\qquad
 B^{\rm rand}_{k,\ell}=B^F_{k,\ell}+\xi_{k,\ell}.
 \end{aligned}
 \label{eq:app-norm-match}
\end{equation}
\begin{equation}
 \begin{aligned}
 R_k^S&=v(x_k^F,t_k,c;B\leftarrow B_k^S)-v_k^0,\\
 R_k^{\rm rand}&=v(x_k^F,t_k,c;B\leftarrow B_k^{\rm rand})-v_k^0.
 \end{aligned}
 \label{eq:app-corrections}
\end{equation}
\begin{equation}
 \begin{aligned}
 c_k^q&=\frac{\langle-J_k^{(\delta)},R_k^q\rangle}
 {\|J_k^{(\delta)}\|_2\,\|R_k^q\|_2},\quad q\in\{S,\mathrm{rand}\},\\
 \Delta\cos_k&=c_k^S-c_k^{\rm rand}.
 \end{aligned}
 \label{eq:app-alignment}
\end{equation}
Both corrections replace all tested blocks at the same $x_k^F$; cosines use flattened predictions and the reduced-feedback direction $-J$. Reverse checks differentiate at native successful states, anchor the random control to $B^S$, and match $\|B^F-B^S\|_F$, without post-hoc sign reversal:
\begin{equation}
 R_k^F=v(x_k^S,t_k,c;B\leftarrow B_k^F)-v(x_k^S,t_k,c;1).
 \label{eq:app-reverse}
\end{equation}
\begin{equation}
 \overline{\Delta\cos}_i(\mathcal W)=\frac{1}{|\mathcal W|}\sum_{k\in\mathcal W}
 \Delta\cos_{i,k},\qquad
 \overline{\Delta\cos}_m(\mathcal W)=\frac{1}{N_m}\sum_{i\in\mathcal I_m}\overline{\Delta\cos}_i(\mathcal W).
 \label{eq:app-pair-aggregation}
\end{equation}
$\mathcal I_m$ contains $N_m$ pairs for model $m$. Average steps within pairs, then pairs within models; Figure~\ref{fig:temporal} weights models equally. Bootstrap and sign tests use pairs. For post-start analysis, obtain the start only from independent suffix transplants:
\begin{equation}
 a_i^*=\max\{a:E_i(I_i^{\{a,\ldots,K\}})=1\},
 \label{eq:app-post-start}
\end{equation}
Use $k\geq a_i^*$; no successful suffix means no start. Check baseline equivalence for recording, $\lambda=0$, successful self-injection, and $\alpha=1$, plus cache alignment, unchanged image-target outputs, matched norms, and finite values.
\subsection{Two-sided sign check}
\label{app:sign}

Figure~\ref{fig:signcheck} in the main text tests whether the same-state sign survives finite interventions and changes final outcomes. Both directions are run at matched strengths on SD3.5M ($\mathrm{cfg}=1$; GenEval: 553 prompts with six seeds per prompt, totaling 3,318 images; base $0.3481$), using \sfgx{} in all joint-attention blocks and with the first-$20\%$ schedule. All twelve feedback-suppression settings ($u<0$ in the auxiliary branch) beat the base; the opposite-sign settings ($u>0$) amplify feedback in the net guidance effect and fall below the base, with lower GenEval scores as $\omega$ grows.

\begingroup
\raggedbottom
\setlength{\abovecaptionskip}{4pt}
\section{Single- and two-object omission probes}
\label{app:object-probes}

This appendix tests whether image-to-text feedback preserves existing errors and whether suppressing it helps generation recover. We use object-omission probes on single- and two-object prompts to track error signals in the evolving text stream and compare them with final-image outcomes. Without control, warnings tend to persist; with \sfgx{}, their persistence falls and many warned failures recover. Suppression can also disrupt initially successful outputs, so we measure both effects in the paired cohort: $72.0\%$ of failures are rescued, whereas $16.7\%$ of successes are disrupted.

\makeatletter
\newcommand{\probecaptype}[1]{\def\@captype{#1}}
\makeatother
\subsection{Probe protocol and paired outcomes}
\label{app:probe-protocol}
\label{app:probe-images}
\label{app:probe-persistence}
We test whether object omissions persist and how I2T feedback suppression changes probe readouts and final images. We reuse OSI's released SD3.5M per-head linear classifiers~\citep{baek2026osi}, without retraining: read normalized, projected text keys for noun T5 tokens after the first 77 CLIP tokens, average noun subtokens, then take the median omission probability over the top 100 heads by released accuracy. They span 14 layers (zero-based 9--20, 22, 23). Track each concept separately and probe guided paths on the clean branch at their current latent.

All runs use SD3.5M, CFG 1.0, 40 steps, $1024\times1024$, and \sfgx{}, with $u_X=-0.4,\omega=8$ in all joint blocks. The 500-pair run enables steps 1--40 and uses official GenEval Mask2Former evaluation. Its 80 prompts yield 233 single-/267 two-object success/failure pairs and 233/306 concept trajectories (539 total); both seeds receive baseline and guided rollouts. These are offline mechanism diagnostics with fixed schedules, not warning-triggered guidance.

A strict symptom exceeds the step-specific 75th percentile of baseline-success readouts and the paired success readout by at least 0.05. Three consecutive symptomatic steps confirm a warning; onset/confirmation are the first/third step, with medians 8/10. By step 20, onset covers 315/539 (58.4\%), but confirmation covers 303/539 (56.2\%). This is paired coverage, not held-out detection accuracy.
\begin{table}[H]
\centering\small
\setlength{\tabcolsep}{4pt}
\caption{\textbf{Omission warnings, generation outcomes, and probe states on 500 SD3.5M trajectory pairs.} Single and Two denote single- and two-object prompts; All combines both. Guided uses SFG-X throughout 40 sampling steps at $\mathrm{cfg}=1$. Count-based rows report events over the eligible concepts or paths, with percentages in parentheses. Symptom persistence and terminal non-recovery are percentages averaged with equal prompt weights.}
\label{tab:probe-main500}
\label{tab:probe-warning500}
\label{tab:probe-rescue500}
\label{tab:probe-interference500}
\label{tab:probe-warned_failure500}
\label{tab:probe-persistence500}
\label{tab:probe-nonrecovery500}
\label{tab:probe-warned_rescue500}
\begin{tabular}{llrrr}\toprule
Metric & Condition & Single & Two & All\\\midrule
Warning coverage (concepts) & Any step & 166/233~(71.2\%) & 224/306~(73.2\%) & 390/539~(72.4\%)\\
 & By step 20 & 118/233~(50.6\%) & 185/306~(60.5\%) & 303/539~(56.2\%)\\
\midrule
Failure rescue (paths) & Guided & 188/223~(84.3\%) & 161/262~(61.5\%) & 349/485~(72.0\%)\\
Success disruption & Guided & 13/228~(5.7\%) & 68/258~(26.4\%) & 81/486~(16.7\%)\\
Final failure after warning & Base & 158/166~(95.2\%) & 201/205~(98.0\%) & 359/371~(96.8\%)\\
 & Guided & 31/166~(18.7\%) & 81/205~(39.5\%) & 112/371~(30.2\%)\\
Warned-failure rescue & Guided & 127/158~(80.4\%) & 120/201~(59.7\%) & 247/359~(68.8\%)\\
\midrule
Symptom persistence & Base & 80.3 & 71.2 & 75.4\\
 & Guided & 24.2 & 30.4 & 27.6\\
No terminal recovery & Base & 76.9 & 59.5 & 67.4\\
 & Guided & 37.5 & 32.6 & 34.8\\
\bottomrule\end{tabular}
\end{table}
Final outcomes weight paths equally: re-evaluation retains 485 failures and 486 successes (single/two denominators 223/262 and 228/258). The 390 warned concepts deduplicate to 371 paths (166/205); warned-failure rescue uses only their 359 failures. Success retention is 405/486 (83.3\%), comprising 215/228 (94.3\%) single and 190/258 (73.6\%) two-object paths. Among warnings confirmed by step 20, baseline failure is 280/290 (96.6\%), rescue 188/280 (67.1\%), and guided failure 92/290 (31.7\%).

Probe metrics weight prompts equally: persistence is the symptomatic-step fraction from baseline onset through step 40, including the trigger; terminal recovery requires at least three symptom-free steps after confirmation and no recurrence. Persistence drops 47.8 points (prompt-cluster bootstrap 95\% interval for guided minus baseline: $[-52.9,-42.7]$). This supports persistence and controllable reduction of object omissions, not monotonic amplification or irreversibility.
 
\subsection{Guidance windows: outcomes and probe states}
\label{app:probe-windows}
\label{app:probe-window-symptoms}
The separate 85-pair cohort uses 50 prompts: 59 single-/26 two-object pairs and 90 concept trajectories. All windows share model, parameters, baseline labels, and probe records; all 85 failures and 85 successes are verified. Compare baseline with inclusive windows 1--10, 1--20, 10--25, and 1--40. Lines connect categorical settings, not a continuous duration sweep. S/T/all denominators are 59/26/85 paths for rescue/disruption and 41/25/66 for warned-failure rescue. Bold marks each column's best guided setting.
\par\medskip\noindent\begin{minipage}[t]{0.43\linewidth}\vspace{0pt}\centering\small
\setlength{\tabcolsep}{3pt}\renewcommand{\arraystretch}{1.08}\probecaptype{table}
\caption{\textbf{SFG-X window effects on final images in the 85-pair SD3.5M cohort.} Windows specify inclusive sampling steps. S/T denote single-/two-object prompts; All combines both. Values are percentages. Arrows indicate the preferred direction; bold marks the best guided window per column and metric.}
\label{tab:probe-window_rescue}
\label{tab:probe-window_interference}
\label{tab:probe-window_warned_rescue}
\begin{tabular}{lrrr}\toprule
Window & S & T & All\\
\midrule\multicolumn{4}{l}{\textit{Failure rescue $\uparrow$}}\\
Base & 0.0 & 0.0 & 0.0\\
1--10 & 81.4 & 42.3 & 69.4\\
1--20 & 83.1 & 57.7 & 75.3\\
10--25 & 67.8 & 42.3 & 60.0\\
1--40 & \textbf{86.4} & \textbf{69.2} & \textbf{81.2}\\
\midrule\multicolumn{4}{l}{\textit{Success disruption $\downarrow$}}\\
Base & 0.0 & 0.0 & 0.0\\
1--10 & 11.9 & 26.9 & 16.5\\
1--20 & \textbf{3.4} & 26.9 & \textbf{10.6}\\
10--25 & 10.2 & \textbf{19.2} & 12.9\\
1--40 & 5.1 & 26.9 & 11.8\\
\midrule\multicolumn{4}{l}{\textit{Warned-failure rescue $\uparrow$}}\\
Base & 0.0 & 0.0 & 0.0\\
1--10 & 78.0 & 44.0 & 65.2\\
1--20 & 80.5 & 60.0 & 72.7\\
10--25 & 65.9 & 44.0 & 57.6\\
1--40 & \textbf{85.4} & \textbf{72.0} & \textbf{80.3}\\
\bottomrule\end{tabular}
\end{minipage}\hfill
\begin{minipage}[t]{0.54\linewidth}\vspace{0pt}\centering
\includegraphics[width=\linewidth]{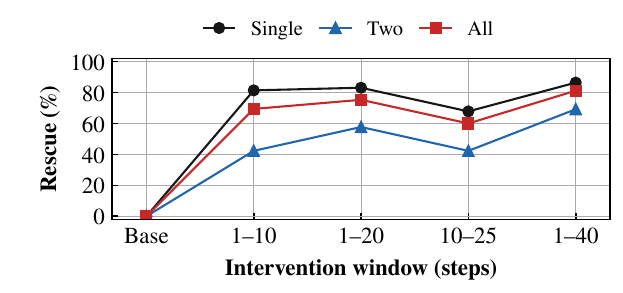}
\includegraphics[width=\linewidth]{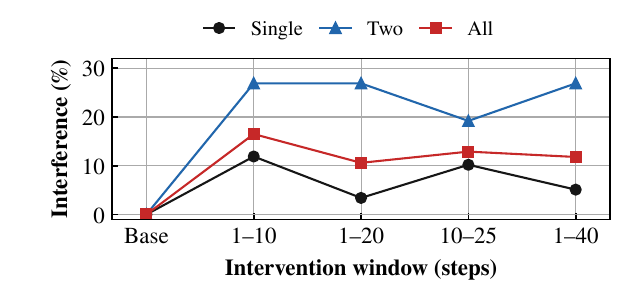}
\includegraphics[width=\linewidth]{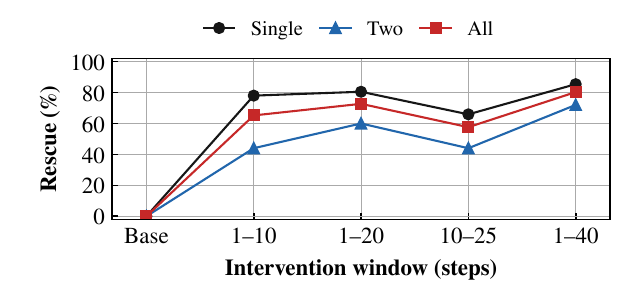}
\probecaptype{figure}
\caption{Top to bottom: failure rescue, success disruption, and warned-failure rescue.}
\label{fig:probe-window_rescue}
\label{fig:probe-window_interference}
\label{fig:probe-window_warned_rescue}
\end{minipage}\par
Full-window rescue is highest (69/85); 1--20 causes the fewest disruptions (9/85). In the above window order, success retention is 71/85 (83.5\%), 76/85 (89.4\%), 74/85 (87.1\%), 75/85 (88.2\%); final failure on the 66 warned paths is 34.8\%, 27.3\%, 42.4\%, 19.7\%. Rescue on the 53 paths confirmed by step 20 is 34/53 (64.2\%), 37/53 (69.8\%), 28/53 (52.8\%), 42/53 (79.2\%). This ordering has not been tested on the 500-pair cohort.
 
Probe-state comparisons fix the same 70 baseline-warned concept trajectories (66 deduplicated failed paths) in the 85-pair cohort, with the same criterion and prompt weighting. Probe recovery and final rescue can rank windows differently: two-object terminal non-recovery is lowest for 1--10, whereas image rescue peaks at 1--40.
\par\medskip\noindent\begin{minipage}[t]{0.43\linewidth}\vspace{0pt}\centering\small
\setlength{\tabcolsep}{3pt}\renewcommand{\arraystretch}{1.08}\probecaptype{table}
\caption{\textbf{SFG-X window effects on omission-probe states in the 85-pair SD3.5M cohort.} S/T denote single-/two-object prompts. Percentages use equal prompt weights on fixed baseline-warned concepts. Lower is better; bold marks the best guided window per column and metric.}
\label{tab:probe-window_persistence}
\label{tab:probe-window_nonrecovery}
\begin{tabular}{lrrr}\toprule
Window & S & T & All\\
\midrule\multicolumn{4}{l}{\textit{Symptom persistence $\downarrow$}}\\
Base & 75.7 & 59.6 & 69.1\\
1--10 & 26.8 & 37.9 & 31.4\\
1--20 & 24.0 & \textbf{25.5} & 24.6\\
10--25 & 44.0 & 34.5 & 40.1\\
1--40 & \textbf{21.3} & 28.2 & \textbf{24.1}\\
\midrule\multicolumn{4}{l}{\textit{No terminal recovery $\downarrow$}}\\
Base & 72.8 & 61.1 & 68.0\\
1--10 & 48.1 & \textbf{29.2} & 40.3\\
1--20 & 48.4 & 37.5 & 43.9\\
10--25 & 51.3 & 35.2 & 44.7\\
1--40 & \textbf{32.7} & 45.8 & \textbf{38.1}\\
\bottomrule\end{tabular}
\end{minipage}\hfill
\begin{minipage}[t]{0.54\linewidth}\vspace{0pt}\centering
\includegraphics[width=\linewidth]{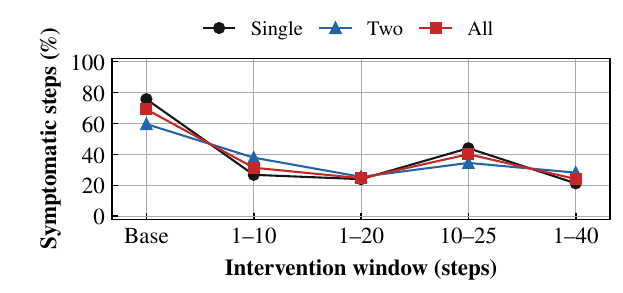}
\includegraphics[width=\linewidth]{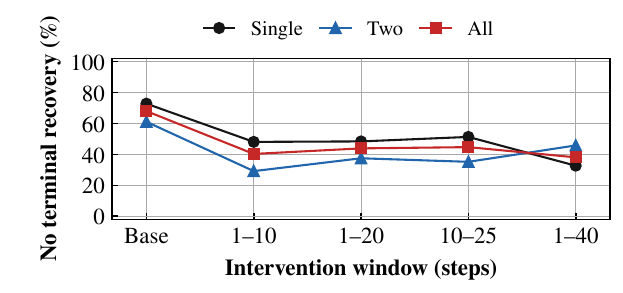}
\probecaptype{figure}
\caption{Top to bottom: symptom persistence and no terminal recovery.}
\label{fig:probe-window_persistence}
\label{fig:probe-window_nonrecovery}
\end{minipage}\par
\subsection{Separate held-out threshold check}
\label{app:probe-heldout}
Use the temporal median over steps 9--24, calibrate on seeds 42/10042, and test seed 20042 with the same released classifiers. The test set has 22 successes and 22 failures for single-object prompts, and 10 each for two-object prompts. q90/q95 are calibration-success percentiles; false alarms flag successful baseline paths and differ from guidance disruption and paired warning coverage.
\begin{table}[H]\centering\small
\caption{\textbf{Omission-probe threshold evaluation on a held-out SD3.5M seed.} S/T denote single-/two-object prompts. q90/q95 are the 90th/95th percentiles of probe scores on successful calibration trajectories. Detect is the percentage of failed test trajectories flagged by the probe; False alarm is the percentage of successful test trajectories flagged.}
\label{tab:probe-heldout}
\begin{tabular}{lrrr}\toprule
Class & Threshold & Detect & False alarm\\\midrule
S & q90 & 40.9 & 18.2\\
S & q95 & 27.3 & 9.1\\
T & q90 & 50.0 & 20.0\\
T & q95 & 20.0 & 10.0\\
\bottomrule\end{tabular}
\end{table}

\endgroup

\section{Qualitative results}
\label{app:qualitative-comparisons}
\begin{figure}[H]
\centering
\includegraphics[width=\linewidth]{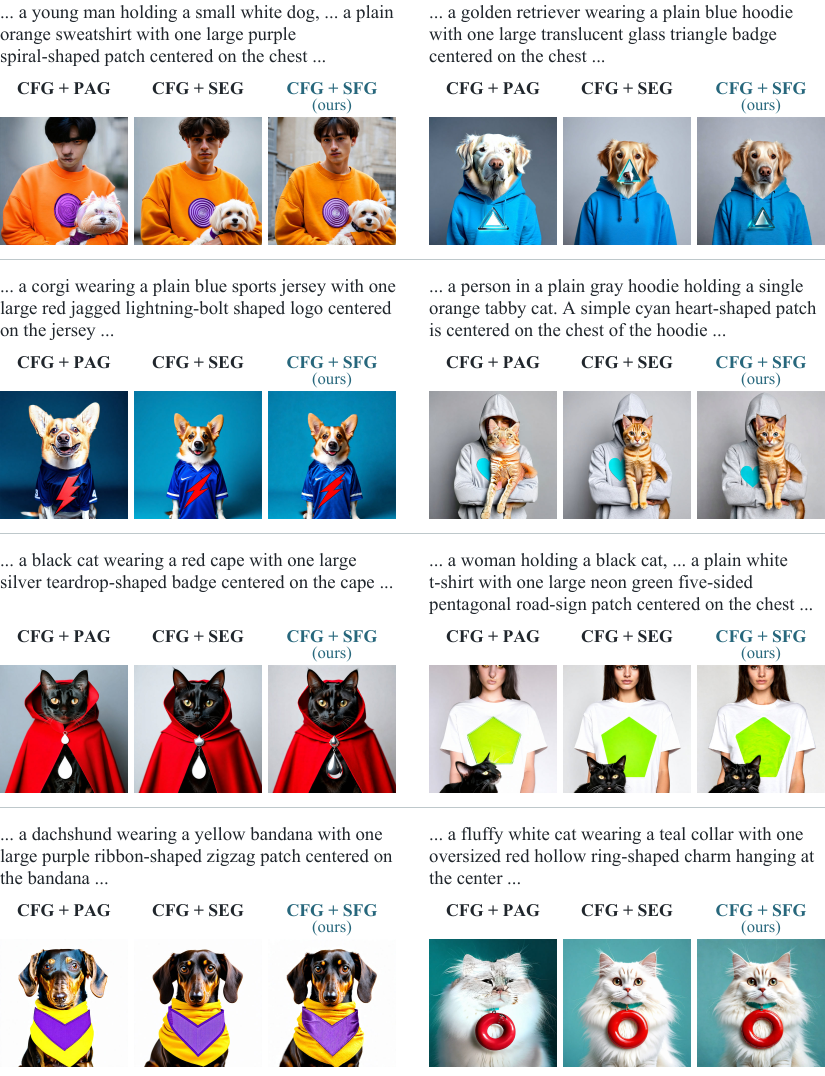}
\caption{\textbf{Qualitative comparisons on SD3.5M.} Each triplet compares CFG+PAG, CFG+SEG, and CFG+SFG (ours) with the same prompt and seed. Prompt excerpts highlight the requested subjects and visual attributes.}
\label{fig:qualitative-overview}
\end{figure}

\begin{figure}[p]
\centering
\includegraphics[width=\linewidth]{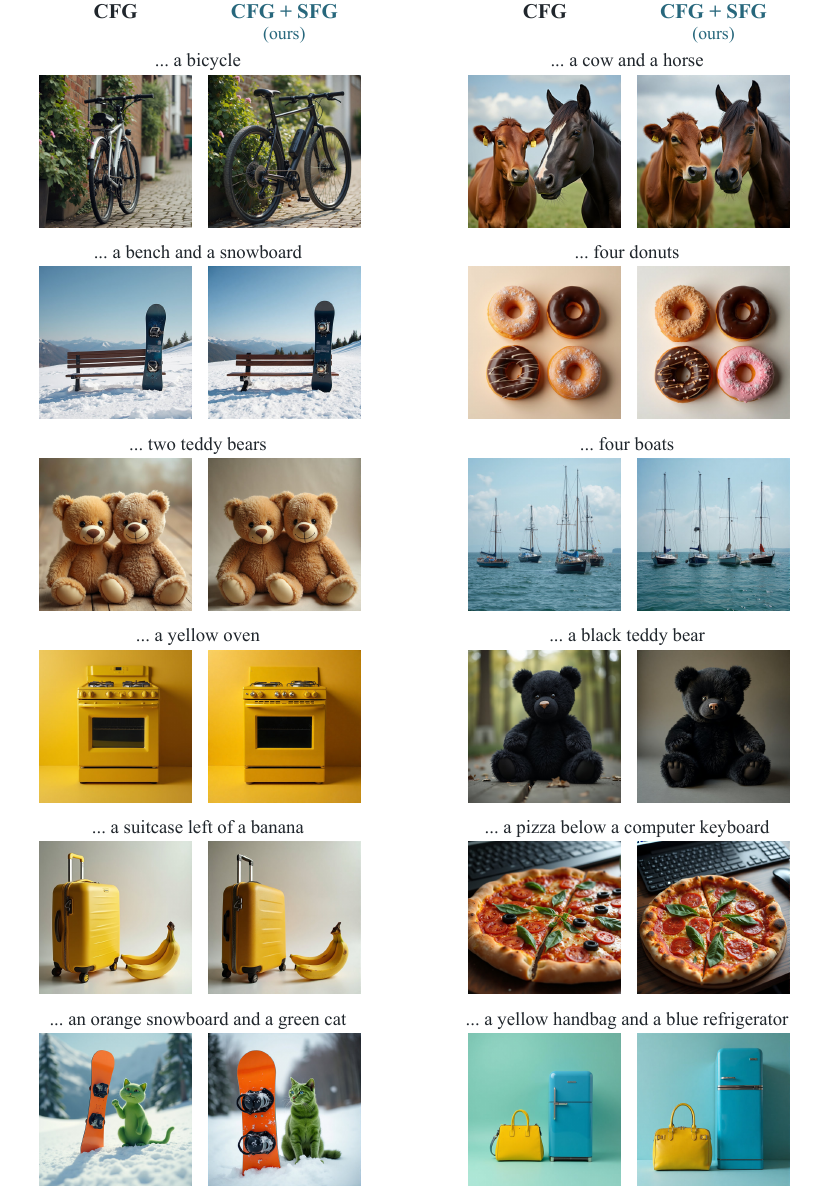}
\caption{\textbf{Compositional comparisons on FLUX.} Each pair compares CFG and CFG+SFG (ours) with the same prompt and seed. Prompt excerpts appear above the pairs.}
\label{fig:qualitative-flux-composition}
\end{figure}

\begin{figure}[p]
\centering
\includegraphics[width=\linewidth]{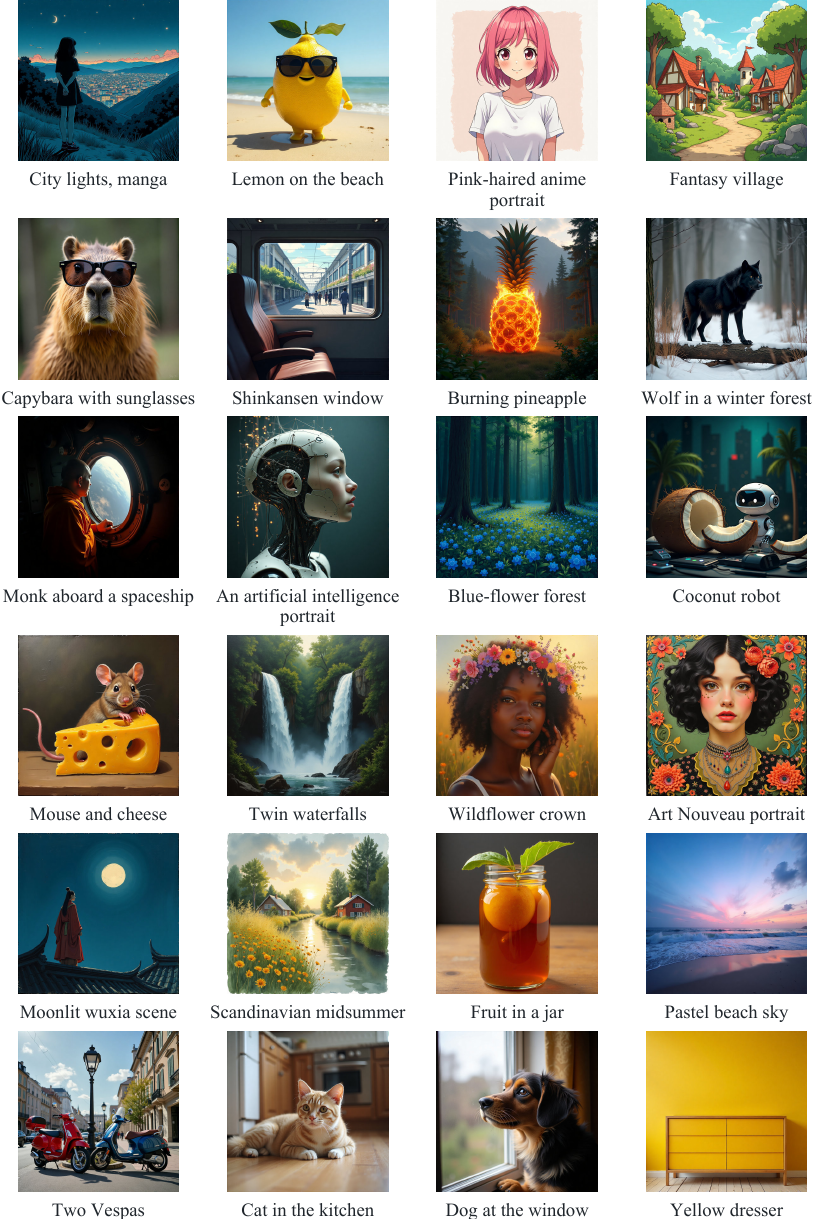}
\caption{\textbf{FLUX with CFG+SFG (ours) across styles and scenes.} Selected outputs from our 800-prompt HPSv2.1 evaluation set; labels describe the images and are not the original prompts.}
\label{fig:qualitative-flux-styles}
\end{figure}

\end{document}